%% file: main.tex
\documentclass{article}

 \usepackage[preprint]{neurips_2026}

\usepackage[utf8]{inputenc} 
\usepackage[T1]{fontenc}    
\usepackage{hyperref}       
\usepackage{url}            
\usepackage{booktabs}       
\usepackage{amsfonts}       
\usepackage{nicefrac}       
\usepackage{microtype}      
\usepackage{xcolor}         
\usepackage{graphicx}
\usepackage{multicol}
\usepackage{multirow}
\usepackage{pifont}
\usepackage{amsmath}
\usepackage{caption}
\usepackage{fontawesome5}
\usepackage{twemojis}
\usepackage{xcolor}
\usepackage{adjustbox}
\usepackage{subcaption}
\usepackage{enumitem}
\usepackage{etoc}
\etocdepthtag.toc{mtchapter}
\etocsettagdepth{mtchapter}{subsection}
\etocsettagdepth{mtappendix}{none}

\definecolor{gold}{RGB}{255,215,0}
\definecolor{silver}{RGB}{192,192,192}
\definecolor{bronze}{RGB}{205,127,50}

\usepackage{xcolor}
\usepackage[table]{xcolor}
\usepackage{hyperref}
\usepackage[most]{tcolorbox}
\definecolor{darkblue}{rgb}{0, 0.40, 0.75}
\definecolor{mygray}{gray}{0.95}
\hypersetup{colorlinks=true, citecolor=darkblue, linkcolor=darkblue, urlcolor=darkblue}
\tcbset{
  aibox/.style={
    width=\linewidth,
    top=8pt,
    bottom=4pt,
    colback=blue!6!white,
    colframe=black,
    colbacktitle=black,
    enhanced,
    center,
    attach boxed title to top left={yshift=-0.1in,xshift=0.15in},
    boxed title style={boxrule=0pt,colframe=white,},
  }
}
\newtcolorbox{AIbox}[2][]{aibox,title=#2,#1}

\usepackage{xcolor}
\usepackage[table]{xcolor}
\usepackage{hyperref}
\usepackage[most]{tcolorbox}

\definecolor{darkblue}{rgb}{0, 0.40, 0.75}
\definecolor{mygray}{gray}{0.95}

\newcommand{\ours}[0]{{VideoASMR-Bench}\xspace}
\usepackage{xspace}

\newcommand{\eg}{\textit{e.g.,}\xspace}

\definecolor{my_yellow}{RGB}{255,165,0}
\definecolor{darkolivegreen}{rgb}{0.33, 0.42, 0.18}
\definecolor{my_green}{RGB}{40,154,121}
\definecolor{my_red}{RGB}{176,46,46}
\newcommand{\correctmark}{\textcolor{my_green}{\ding{52}}} 
\newcommand{\errormark}{\textcolor{my_red}{\ding{56}}}

\definecolor{lightpurple}{RGB}{230, 230, 250}

\title{\ours: Can AI-Generated ASMR Videos Fool VLMs and Humans?}

\author{%
Jiaqi Wang$^{1}$ \thanks{Equal contribution.} \quad
Weijia Wu$^{2*}$ \quad
Yi Zhan$^{3}$ \quad
Rui Zhao$^{2}$ \quad
Ming Hu$^{4}$ \\
James Cheng$^{1}$ \quad
Wei Liu$^{5}$ \quad
Philip Torr$^{6}$ \quad
Kevin Qinghong Lin$^{6}$ \\
\\
$^{1}$The Chinese University of Hong Kong \quad
$^{2}$National University of Singapore \\
$^{3}$Peking University \quad
$^{4}$Monash University \quad
$^{5}$Video Rebirth \quad
$^{6}$University of Oxford \\
\\
Project Page: \url{https://video-reality-test.github.io/} \\
}

\begin{document}

\maketitle

\begin{abstract}
  With AI-generated videos increasingly indistinguishable from reality, current benchmarks primarily focus on broad semantic alignment and basic physical consistency, offering limited discriminative power for evaluating them.
To address this, we introduce \textbf{\ours}, a benchmark based on \textit{Autonomous Sensory Meridian Response (ASMR) videos} that emphasizes fine-grained audio-visual perception and sensory immersion. 
This benchmark aims to answer two key questions:
\textbf{\textit{(i)} Are today’s video understanding models (VLMs) sensitive enough to detect AI-generated ASMR videos by recognizing minor visual, physical, or auditory artifacts?}
\textbf{\textit{(ii)} Can today’s video generation models (VGMs) produce convincing ASMR videos with immersive experiences?} 
This benchmark comprises a diverse set of 1,500 high-quality real ASMR videos curated from social media, alongside 2,235 synthetic counterparts generated by nine VGMs. 
Additionally, we open-source an extensible suite of prompts and reference images, enabling the benchmark to scale dynamically with future video models. 
Moreover, we design an automatic understanding–generation evaluation framework between VGMs and VLMs, where VGMs aim to produce realistic fake videos 
to fool the VLMs, while the VLMs seek to detect them, forming an adversarial game between the two parties. 
 Our evaluation on \ours reveals that even state-of-the-art VLMs, such as Gemini-3-Pro, fail to reliably detect AI-generated ASMR videos.
Meanwhile, current frontier video generation models can produce indistinguishable ASMR videos for VLMs, while humans can still identify them relatively easily.

\end{abstract}

\section{Introduction}
The boundary between real and synthetic videos blurs with the dramatic advent of frontier video generation models (VGMs) such as Sora2 \cite{sora2}, Veo3 \cite{veo3}. 
These VGMs have moved beyond generating short, low-resolution clips to to producing visually coherent, high-definition videos that are increasingly indistinguishable from reality~\cite{ma2025step,huang2025step,kong2024hunyuanvideo,wan2025wan}. 
Despite this progress, current evaluation suites~\cite{huang2024vbench,bansal2024videophy,phygenbench,genvidbench,ye2024loki,gen_video,bai2025impossible} still focus on semantic and physical perspectives, which offer limited discriminative power for assessing modern AI-generated videos.
This motivates a fine-grained, high-sensitivity assessment to guide the next stage of video perception and generation research.

\begin{figure}
    \centering
    \includegraphics[width=\linewidth]{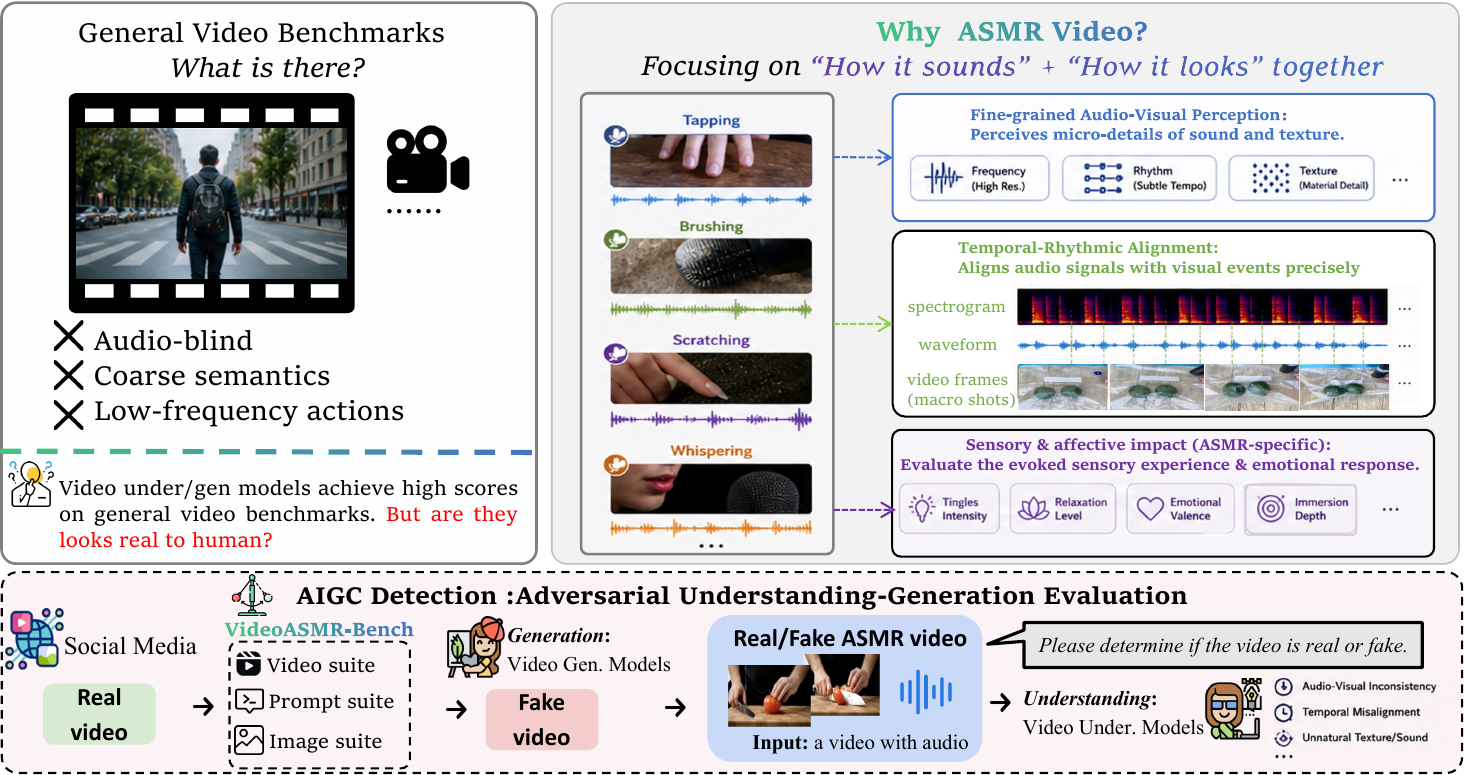}
    \caption{\textbf{(Top) Motivation of why we study ASMR Video?} General video benchmarks are high-fidelity audio, macro videography, and subtle textures, thus insufficient for capturing the fine-grained audio-visual understanding contexts in ASMR videos. 
    \textbf{(Bottom) Illustration of \ours} which introduces a multi-level taxonomy spanning perception, temporal alignment, and sensory-affective impact. We pioneer the task of AIGC detection in the ASMR domain, aiming to paving the way for next-generation video models.}
    \label{fig:teaser}
    \vspace{-1em}
\end{figure}

To address this, we propose \ours (Fig.~\ref{fig:teaser}), a benchmark designed to assess sensory immersion through the lens of \textbf{Autonomous Sensory Meridian Response (ASMR)}.
ASMR videos (\eg tapping, brushing, and whispering) pose three uniquely challenging evaluation dimensions: fine-grained audio-visual perception, precise temporal-rhythmic alignment, and sensory-affective impact, making them an ideal stress test for video generation and understanding models. 
To the best of our knowledge, this is the first work to systematically study both video generation and understanding in the ASMR domain.
Based on this benchmark, we aim to answer two key questions: 
\textit{(i)} Are today’s video understanding models (VLMs) sensitive enough to detect AI-generated ASMR videos by recognizing minor visual, physical, or auditory artifacts?
\textit{(ii)} Can today’s video generation models (VGMs) produce convincing ASMR videos with tightly coupled audio and visual signals?

We start by collecting a large and diverse set of popular ASMR videos from social media, totaling 1,500 high-quality clips.
Based on these real ASMR videos, we construct the ASMR Prompt and Image Suite to synthesize 2,235 fake ASMR videos across 9 state-of-the-art VGMs across 4 different settings. 
We categorize them into easy and hard levels: easy clips feature short, single-step actions in uniform scenes, whereas hard clips involve longer, multi-step activities with diverse materials in complex environments.
Notably, we formalize an automatically adversarial understanding-generation evaluation between VGMs and VLMs, where VGMs aim to produce fake videos as real as possible to fool the VLMs, while the VLMs seek to detect as many fake videos as possible, forming an adversarial game between the two parties. 

We conduct a comprehensive evaluation of eleven VLMs and nine VGMs, revealing several notable results. 
First, \textbf{even state-of-the-art VLMs such as Gemini-3-pro struggle to detect AI-generated ASMR videos}, lagging significantly behind humans, with a detection score of 45 compared to a human score of 70. 
Second, \textbf{current video generation models produce ASMR videos that are largely indistinguishable to VLMs}; the best-performing model, Veo3.1, had only 12.54\% of its generated videos correctly identified as fake, while humans can still identify them relatively easily.
Finally, we identify several key observations and insights. For example, VLMs exhibit fundamental limitations in ASMR detection: (i) Watermarks act as detection shortcuts causing an average performance drop of $\sim$30 points. 
(ii) VLMs show a strong bias toward classifying videos as real, misclassifying up to approximately 71\% of fake videos.
(iii) Auditory cues, though largely overlooked, significantly enhance discriminative capability. 
To mitigate these limitations, our ASMR prompt suite consistently improves realism scores across models.
Moreover, we find that preference-based evaluation outperforms direct judgment, and explicit reasoning traces further enhance detection accuracy.
In summary, our contributions include:

\begin{itemize}[leftmargin=*]
    \item To the best of our knowledge, this is the first work to investigate the ASMR domain, which explores the immersive sensory experience for advanced video understanding and generation research.
\item 
\ours is sourced from popular social media and consists of a high-quality ASMR video suite covering diverse sceneries, an ASMR prompt suite, and an ASMR image suite, supporting our benchmark to scale by conditional synthesis of additional fake videos.
\item 
We design an adversarial understanding–generation evaluation framework spanning 11 VLMs and 9 VGMs, showing that VLMs remain unreliable for detecting AI-generated ASMR videos, while VGMs produce indistinguishable sensory-immersive videos for VLMs.
\end{itemize}

\input{table/compare}

\section{Related Work}
This work evaluates video models across two key domains: video understanding and video generation. Tab.~\ref{tab:bench comparasion} presents a comprehensive comparison of \ours and existing video benchmarks, highlighting their relationships and key differences. We then outline the primary objectives of existing models and benchmarks in each domain.

\noindent\textbf{Video Generation and Understanding Benchmarks.}
Current benchmarks primarily evaluate visual and semantic quality. For generation, existing suites like VBench~\citep{huang2024vbench} and T2V-CompBench~\citep{sun2024t2v} primarily assess visual quality, temporal consistency, and text-video alignment. While recent works like VideoPhy~\citep{bansal2024videophy} and PhyGenBench~\citep{meng2024towards} have begun to evaluate physical correctness, they focus on gross violations (e.g., object disappearance, gravity) rather than fine-grained material interactions.
For understanding, benchmarks like MVBench~\citep{li2024mvbench} and Video-MME~\citep{video_mme} target high-level semantic tasks like captioning or QA, often treating audio as a secondary modality or ignoring it entirely.
Consequently, these disparate evaluations fail to capture the immersive sensory realism required by frontier models. 
In contrast, our benchmark \ours adopts ASMR as a stringent testbed for fine-grained physical realism and proposes an adversarial framework that jointly evaluates generation and perceptual understanding.

\noindent\textbf{Joint Audio-Visual Realism and Detection.}
While recent models like MM-Diffusion~\citep{ruan2023mm} and AV-DiT~\citep{wang2024av} advance joint video-audio synthesis, evaluations often settle for coarse synchronization rather than high-fidelity coupling. 
Furthermore, existing AI-generated video detection benchmarks~\citep{gen_video,bai2025impossible} predominantly rely on visual-only cues, ignoring the critical role of audio in verifying authenticity. 
Omni-modal models~\citep{qwen25omni} integrate audio but lack dedicated protocols for detecting synthetic content. 
To address this, \ours is the first benchmark that explicitly uses strict audio-visual alignment and sensory immersion as the core criteria to distinguish real from fake.

\section{\ours}
Our benchmark, \ours, comprises three core components: the ASMR Video Suite, the ASMR Image Suite, and the ASMR Prompt Suite, all curated following the pipeline introduced in Sec.~\ref{real data}. 
The ASMR Image Suite and ASMR Prompt Suite further serve as the foundation for generating synthetic ASMR videos to extend the scale of \ours, as detailed in Sec.~\ref{fake data}. 
We then present a comprehensive data analysis in Sec.~\ref{sec: examples}, including statistics across multiple dimensions. Finally, we formalize a novel evaluation protocol in Sec.~\ref{sec: eval}.

\begin{figure*}[!t]
    \centering
    \includegraphics[width=1\linewidth]{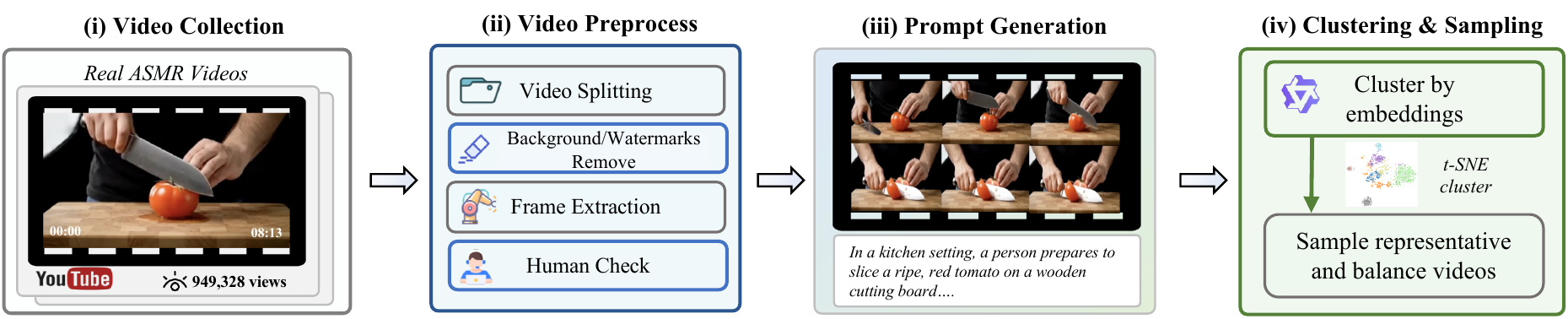}
    \caption{ 
    \textbf{Four phases of ASMR data collection} in \ours, resulting in the ASMR Video, Prompt and Image Suites.
    }
    \vspace{-1em}
    \label{fig1:benchmark pipeline}
\end{figure*}

\subsection{Collect ASMR Suite Families}
\label{real data}

\textbf{Overview.} As illustrated in Fig.~\ref{fig1:benchmark pipeline}, \ours consists of ASMR Video Suite, ASMR Image Suite, and ASMR Prompt Suite. The ASMR Image Suite and ASMR Prompt Suite facilitate the synthesis of fake ASMR videos via video generation models, scaling our benchmark.

\begin{itemize}[leftmargin=*]
    \item \textit{\textbf{ASMR Video Suite}} comprises 1,500 sensory-immersive ASMR videos collected from social media, organized into easy and hard levels, covering six different dimensions. These videos emphasize precise audio–visual synchronization, high-fidelity physical interactions across diverse materials, and sustained sensory immersion in varied scenes.
    \item \textit{\textbf{ASMR Prompt Suite}} incorporates the ASMR text description for each ASMR video. 
The text prompt describes environments, main subjects or objects, actions, temporal dynamics, and corresponding sounds.These prompts act as conditioning signals to scaling the size of our benchmark across successive iterations of various VGMs.

\item \textit{\textbf{ASMR Image Suite}} incorporates matched high-quality referred ASMR images.
As a natural starting point, for each ASMR video we use the first frame as conditioning for generation, optionally combined with the ASMR prompt suite, to create diverse synthetic fake ASMR videos.
\end{itemize}

\vspace{-2mm}
\textbf{Construction.} As seen in Fig~\ref{fig1:benchmark pipeline}, we propose a comprehensive framework for developing \ours through the following four steps to ensure the quality of the real ASMR data collection:

\textbf{(i) Raw Video Collection.}
As seen in Fig~\ref{fig1:benchmark pipeline}(a), to curate high-quality and widely engaging ASMR data, we collect ASMR videos from the social media platform YouTube, using view count as a proxy for human immersion and preference, and retain those exceeding 900K views. Following this criterion, we gather over 100 raw ASMR videos, totaling more than 2,000 minutes. 

\textbf{(ii) Video Preprocess.}
Then we apply four steps to filter and clean these raw video, resulting in 1,500 high-quality clips.
As seen in Fig~\ref{fig1:benchmark pipeline}(b): (a) We split short-video compilations into individual clips.
(b) We remove artificial backgrounds and watermarks to reduce visual interference.
(c) For each processed clip, we extract the first video frame to construct the ASMR Image Suite.
(d) All steps are manually inspected to ensure that the ASMR videos are correctly segmented and that text cues and watermarks are successfully removed.

\textbf{(iii) ASMR Prompt Suite Generation.} 
As seen in Fig~\ref{fig1:benchmark pipeline}(c), we divide the data into 735 easy levels, primarily consisting of kitchen scenes involving food cutting, and 1,500 hard levels where environments, actions, and objects vary more substantially.
For the easy level, we use Sora2’s storyboard module to generate prompt conditioned on the ASMR image suite. 
For the hard levels, which are larger in scale, we leverage Gemini-2.5-Pro via its API to generate prompt based on eight uniformly sampled frames per video, with details provided in App.~\ref{app: different prompt}.
All prompts are manually inspected by human annotators.


\textbf{(iv) Clustering \& Sampling.}
As seen in Fig~\ref{fig1:benchmark pipeline}(d), to improve efficiency and facilitate subsequent video generation, we cluster the large scale of the videos into eight semantic classes by applying Qwen3-Embedding-4B for the ASMR prompt suite (Visualization in Fig.~\ref{fig:data1}(a)).
To mitigate class imbalance, we uniformly sample instances from each cluster as representative inputs for downstream video generation.
Finally, both the prompts and the quality of the generated videos are evaluated via human assessment.

\subsection{Synthetic Fake ASMR Videos }
\label{fake data}

\textbf{Different Video Generation Models.} 
We use six VGMs to produce a diverse set of fake ASMR videos, including the latest open-source models (Wan2.2; Opensora-V2; HuyuanVideo) and closed-source models (Sora2, Veo3.1-fast).
We use the default text-image-to-video generation settings, using our ASMR prompt suite and ASMR image suite.
This process produces a total of 894 synthetic videos.
See App.~\ref{sec: video generation settings} for implementation details.

\textbf{Different Generation Settings.} 
To increase the scale and diversity of the benchmark, we further explore ASMR videos generated under multiple settings. 
Besides the default text-image-to-video, we try three additional settings: 
(i) text-to-image-to-video generation, where the model first converts ASMR prompts into images and then generates videos; 
(ii) text-only generation without images,
and (iii) image-only generation with a minimal prompt (“Generate the video as real as possible”). 
This process yields 1,192 synthetic videos in total.
See App.~\ref{sec: video generation settings} for detailed generation settings.

\begin{figure*}[t]
    \centering
    \includegraphics[width=\linewidth]{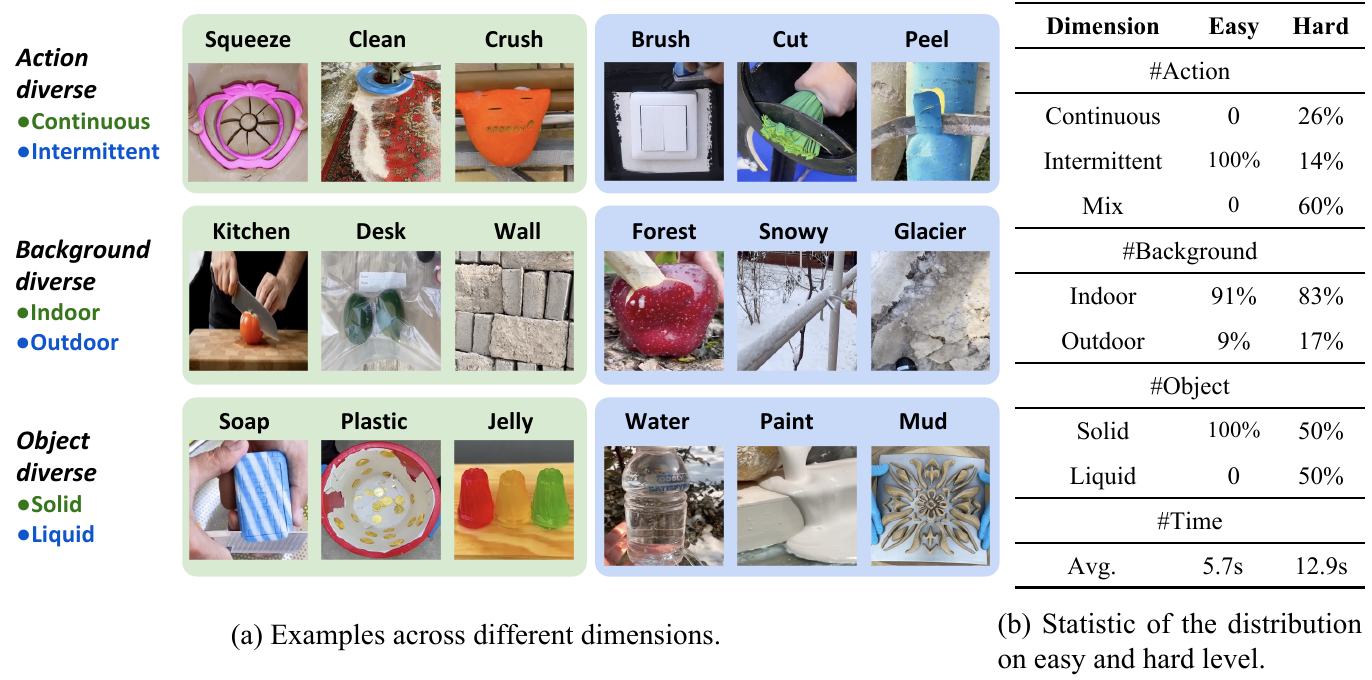}
    \vspace{-1.5em}
    \caption{\textbf{Comprehensive Scope of \ours.} 
    \textbf{(a)} is the example of \ours across different dimensions, \textbf{(b)} is the statistic of the distribution on easy and hard levels.}
    \vspace{-1em}
    \label{fig:datadist}
\end{figure*}

\begin{figure}
    \centering
    \includegraphics[width=\linewidth]{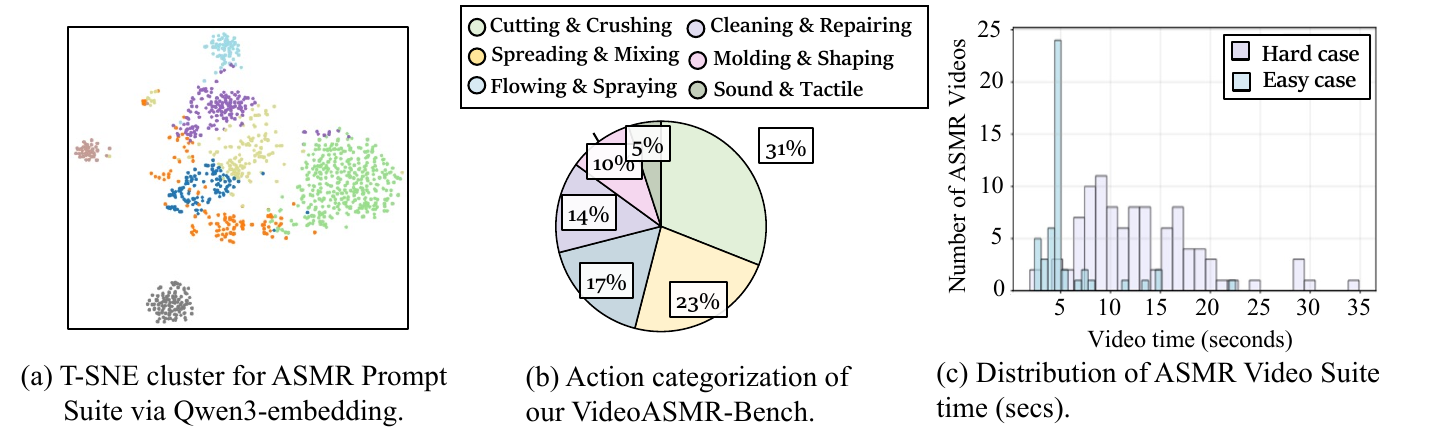}
    \vspace{-1em}
    \caption{\textbf{Detailed analysis of \ours.} \textbf{(a)} is a visualization of a cluster of ASMR prompt suite, \textbf{(b)} is the action statistics of \ours, \textbf{(c)} is the video time distribution and comparison on easy and hard levels.}
    \vspace{-1em}
    \label{fig:data1}
\end{figure}

\subsection{Quality Verification}
\label{sec: examples}
\textbf{Data Statistic.} \ours~contains 2,235 videos forming real–fake ASMR pairs, divided into \textit{easy} (735 videos) and \textit{hard} (1,500 videos) subsets across nine different VGMs under four different video generation settings. Unlike prior benchmarks with a fixed set of pre-generated fake samples, we emphasize both quality and scalability.
Beyond the curated videos in the paper, we will publicly release the whole ASMR prompt suite and image suite that are manually verified to ensure high quality.
Importantly, the benchmark is designed to be extensible: newly released video generation models can be continuously incorporated, allowing the dataset to grow over time while preserving evaluation consistency.

\textbf{ASMR Videos Taxonomy \& Diversity.}
Our \ours videos vary across actions, backgrounds, and objects, with corresponding statistics provided in the Fig.~\ref{fig:datadist}.
In particular, Fig.~\ref{fig:data1}(b) shows that actions are distributed across categories, with Cutting \& Crushing (31\%) and Spreading \& Mixing (23\%) dominating.
Our ASMR videos cover diverse times as seen in Fig.~\ref{fig:data1}(c), where easy videos are typically short ($3\text{--}5$ seconds, while hard videos are substantially longer, up to 20+ seconds.

\textbf{Human Annotation.} For each video generation model, we sample an equal number of generated and real ASMR videos. We recruit 10 annotators with diverse backgrounds, including video generation researchers and non-experts, as well as PhD students and undergraduates. We provide no prior information about the real-to-fake ratio. Audio was included whenever available. Each annotator makes independent judgments on whether each video was real or generated. 



\begin{figure*}[!t]
 \centering
    \includegraphics[width=\linewidth]{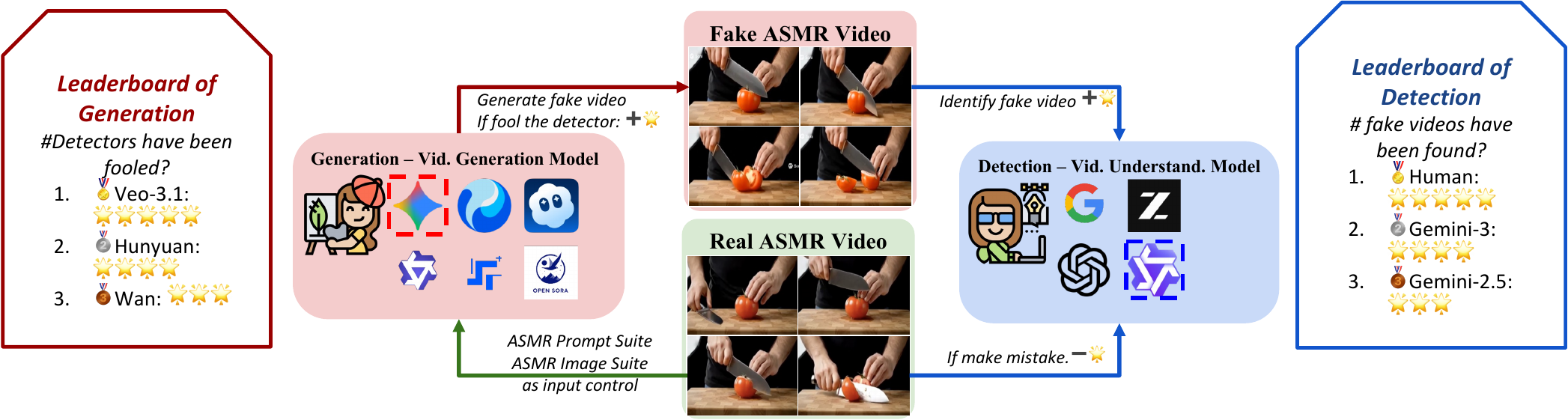}
    \caption{\textbf{An overview of adversarial understanding-generation evaluation.} 
    VGMs synthesize fake ASMR videos that can fool VLMs, while VLMs detect fakes. Leaderboards on both sides highlight which VGMs deceive the most VLMs and which VLMs identify the most fake videos, revealing an adversarial process between generation and understanding.}
    \label{fig:overview}
    \vspace{-1em}
\end{figure*}

\subsection{Adversarial Understanding-Generation Evaluation}
\label{sec: eval}

As illustrated in Fig.~\ref{fig:overview}, we formalize an adversarial game between \textit{video generation models} (VGMs) and \textit{video understanding models} (VLMs). VGMs aim to produce fake videos indistinguishable from real ones, while VLMs seek to detect them — forming a dynamic adversarial interaction.

\textbf{VLM Detection Accuracy.}
VLMs are evaluated on a balanced set of real and fake videos $\mathcal{V}^{g}_{\text{full}} = \{\mathcal{V}_{\text{real}}\} \cup \{\mathcal{V}^{g}_{\text{fake}}\}$ (1:1 ratio), where $\mathcal{V}_{\text{real}}$ denotes real ASMR videos and $\mathcal{V}^{g}_{\text{fake}}$ denotes Fake ASMR videos generated by VGM $g$. Each video is classified as real~($y=1$) or fake~($y=0$). Detection accuracy for a VLM $\mathcal{U}$ against a specific VGM $g$, and its aggregate across all VGMs $\mathcal{G}$, are:
\[
\text{Acc}^{g}_{\mathcal{U}}
= \frac{1}{|\mathcal{V}^{g}_{\text{full}}|}
\sum_{\mathcal{V} \in \mathcal{V}^{g}_{\text{full}}}
\mathbf{1}\!\left[\mathcal{U}(\mathcal{V}) = y(\mathcal{V})\right],
\qquad
\text{Acc}_{\mathcal{U}}
= \frac{1}{|\mathcal{G}|}
\sum_{g \in \mathcal{G}}
\text{Acc}^{g}_{\mathcal{U}}.
\]

\textbf{VGM Realism Score.}
Unlike VLMs, VGMs are evaluated solely on their own generated fake videos $\mathcal{V}^{g}_{\text{fake}}$ — specifically, the proportion that successfully fool VLMs. Here $\mathbf{1}[\mathcal{U}(\mathcal{V})=0]$ counts videos that VLM $\mathcal{U}$ incorrectly classifies as fake. This rate, aggregated across all VLMs $\mathcal{U}$, is:
\[
\text{Fakeness}^{u}_{\mathcal{G}}
= \frac{1}{|\mathcal{V}^{g}_{\text{fake}}|}
\sum_{\mathcal{V} \in \mathcal{V}^{g}_{\text{fake}}}
\mathbf{1}\!\left[\mathcal{U}(\mathcal{V}) = 0\right],
\qquad
\text{Fakeness}_{\mathcal{G}}
= \frac{1}{|\mathcal{U}|}
\sum_{u \in \mathcal{U}}
\text{Fakeness}^{u}_{\mathcal{G}}.
\]

A \textit{higher} $\text{Acc}_{\mathcal{U}}$ indicates a stronger VLM; a \textit{lower} $\text{Fakeness}_{\mathcal{G}}$ indicates more realistic VGM outputs. Crucially, we do not treat this as a simple binary classification problem. Models are explicitly required to produce a \textbf{step-by-step reasoning process} before outputting their final answer. 
In Sec.~\ref{sec: qualitative results}, we analyze the reasoning chains in depth and find that they comprehensively capture fine-grained perceptual cues—such as physical plausibility, audio-visual synchronization, and material texture—rather than relying on superficial shortcuts.
Detailed definitions and metrics are in App.~\ref{app: evaluation metric}.



\section{Evaluation for ASMR Video Understanding}
\label{sub: eval vid un}

\input{table/tabl1}

In this section, we conduct experiments on various video understanding models to evaluate their abilities based on our \ours. 

\textbf{Setup and Evaluation.} We evaluate a range of popular video understanding models (VLMs), encompassing
both latest open-source and proprietary models.
Real ASMR videos are from \ours. 
All fake ASMR videos generated by video generation models are under the ImgText2Vid setting, using our ASMR prompt suite and ASMR image suite. For models that support only text input, we use the ASMR prompt suite only.
Following~\cite{bai2025impossible}, we uniformly sample eight interleaved frames from each initial video as input for video understanding. 


\textbf{Current VLMs fail for detecting the AI-generated ASMR videos.}
Table~\ref{tab:table2} reports the detection accuracy and F1 scores of 10 representative VLMs on \ours across 4 state-of-the-art VGMs, where the inputs include video and audio streams. Human and random baselines are provided as upper and lower bounds, respectively.
We include both open-source and closed-source models for a comprehensive comparison.
Overall, current VLMs perform worse than humans, with most models achieving accuracy below 70 (compared to 87 for humans) and F1 scores below 45 (compared to 70 for humans). Specifically, proprietary VLMs show stronger accuracy than most open-source models, with Gemini-3-pro achieving the best results.
However, their F1 scores remain lower than or only comparable to those of open-source models: GLM-4.5V achieves the best average F1 score across all VLMs, indicating substantial room for improvement in current VLMs.

\begin{figure*}[!t]
    \centering
    \includegraphics[width=1\linewidth]{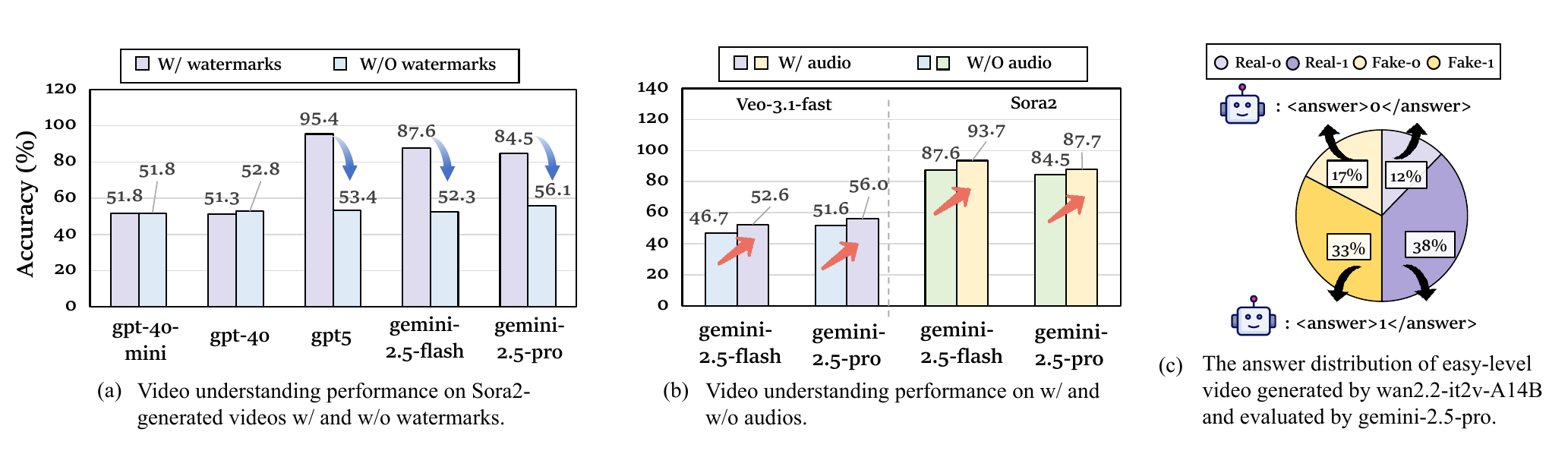}
    \vspace{-2.5em}
    \caption{\textbf{More observations on VLMs.} (a) VLMs' performance drops after the sora watermark removal.
    (b) Incorporating audio along with visual inputs generally improves reality detection accuracy. 
    (c) VLMs have a bias that tends to classify videos as real
rather than fake}
    \vspace{-1em}    
    \label{fig: ablation-1}
\end{figure*}

\section{Evaluation for ASMR Video Generation}
\label{sub: eval vid gen}
\textbf{Setup and Generation.} We evaluate mainstream video generation models,
including the latest open-source 
and closed-source models.
The default video generation setting is ImgText2Vid, using text from the ASMR prompt suite and images from the ASMR image suite. We also explore three additional settings: (i) Text2Img2Vid, where models first convert ASMR prompts into images and then generate videos; (ii) Text2Vid, text-only generation from the ASMR prompt suite; and (iii) Img2Vid, image-only generation with the minimal prompt “Generate the video as real as possible.”

\input{table/opensora_ablation}
\begin{figure}
    \centering
    \includegraphics[width=\linewidth]{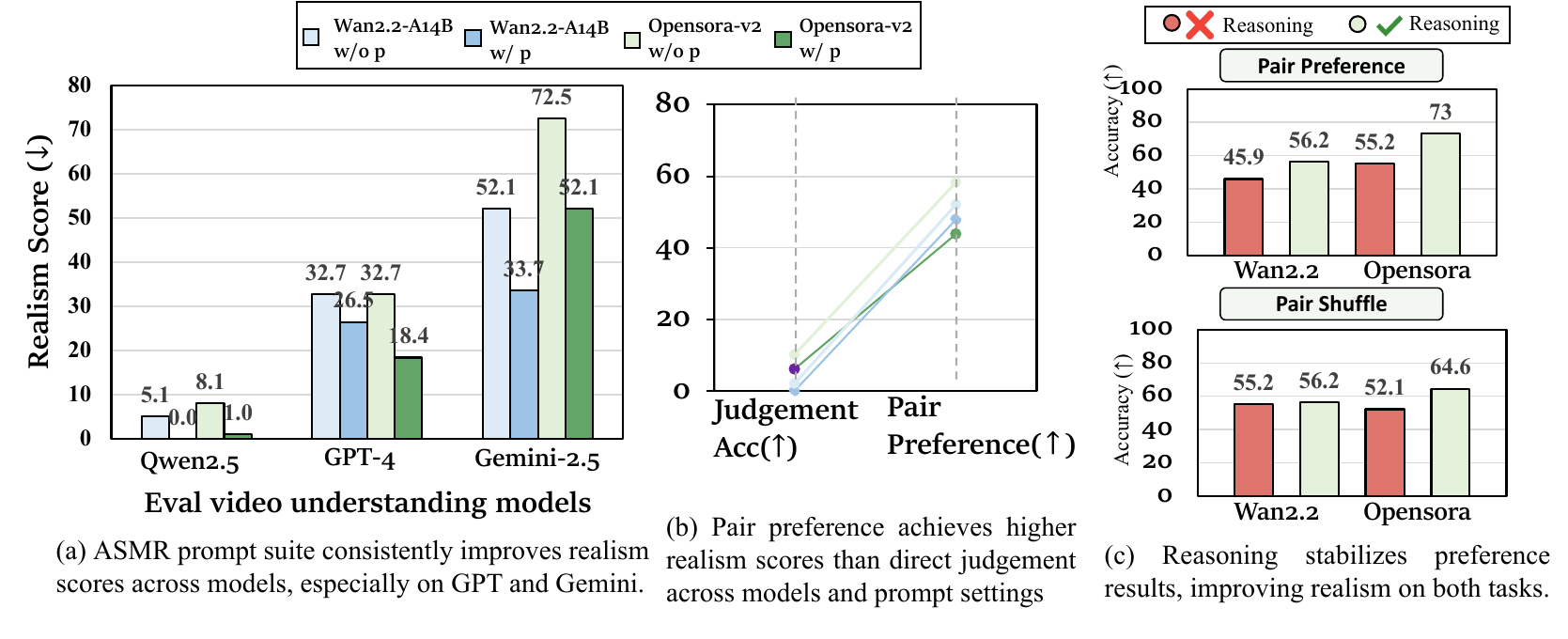}
    \vspace{-2em}
    \caption{Detailed experiment analysis on the easy split of \ours.}
    \label{fig:bench ablation}
    \vspace{-1.5em}
\end{figure}

\textbf{Current video generation models produce indistinguishable ASMR videos for VLMs.}
As shown in Table~\ref{tab:v_gen}, we report the $\text{Fakeness}_{\mathcal{G}}$ for different VGMs under different generation settings.
Overall, the proprietary Veo3.1-Fast performs best, with only 12.54\% of its generated videos identified as fake. 
Notably, the open-source models HunyuanVideo-I2V and Wan2.2-A14B follow closely with detection rates of 14.77\% and 16.10\%, respectively, both surpassing the closed-source Sora2 (without watermark). 
This result suggests that the performance gap between proprietary and open-source video generation models is rapidly narrowing.

\section{Observations and Insights}
\label{sec: obser}

\begin{figure*}[!t]
    \centering
    \includegraphics[width=1\linewidth]{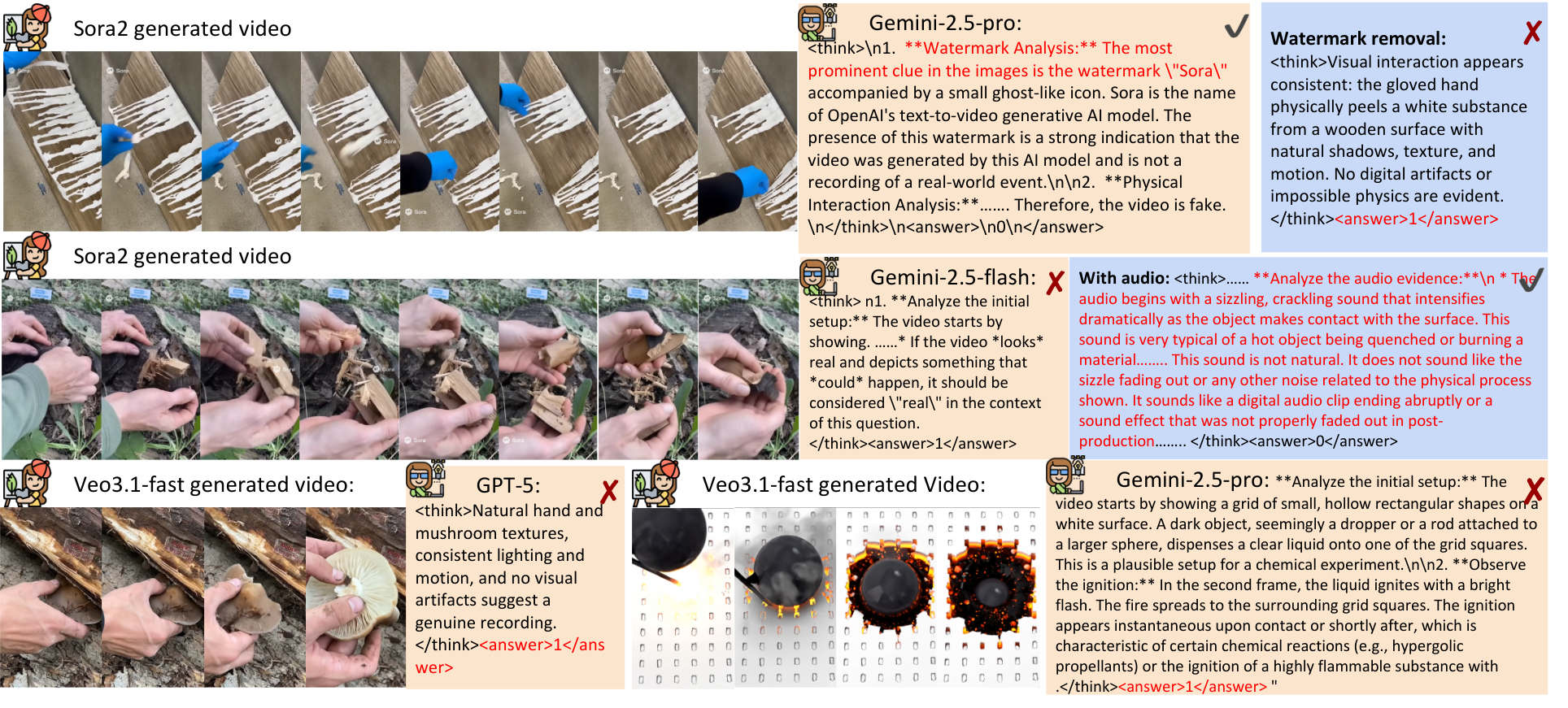}
    \vspace{-2em}    
    \caption{\textbf{Reasoning Traces for Representative Examples.}
    Top: with and without watermarks. 
    Middle: with and without audio cues. 
    Bottom: Veo3.1-Fast generated videos successfully deceive the VLMs. We highlight the incorrect answer and the key thinking process in red.
    }
    \vspace{-1em}
    \label{fig:example}
\end{figure*}

\textbf{Integration of auditory cues enhances the discriminative capabilities of VLMs.}
Fig.~\ref{fig: ablation-1}(b) shows that incorporating audio significantly improves real-versus-AI detection performance, with an average gain of 5 points on the two most realistic fake ASMR videos generated by Veo3.1 and Sora2.
As shown in Fig.~\ref{fig:example}, upon carefully examining the chain-of-thought content, we attribute this improvement to the inadequate audio–visual alignment in current video generation models — a limitation that is particularly harmful for ASMR content, where consistency and immersion are essential.
Interestingly, we find that Sora2 often produces human speech instead of continuous ASMR sounds, whereas Veo3.1-fast generates more action-aligned audio without speech.

\textbf{Watermarks act as the detection shortcut, particularly for strong VLMs.} 
We find that state-of-the-art VLMs often detect the default Sora2 watermark and directly label videos as fake, despite no prompt cues.
As shown in Fig.~\ref{fig: ablation-1}(a), GPT-4o and GPT-4o-mini remain stable, while GPT-5 and Gemini-2.5-Pro/Flash drop sharply (e.g., 95.4 to 53.4), with an average decrease of $\sim$30 points.
This confirms that leading VLMs such as Gemini-2.5 and GPT-5 rely heavily on Sora2 watermarks rather than intrinsic authenticity cues.

\textbf{VLMs have a bias that tends to classify videos as real rather than fake.}  
As shown in Fig.~\ref{fig: ablation-1}(c), we observe a consistent bias toward predicting real across all VLMs.
This trend persists across various VLMs on the easy split of \ours (details in Tab.~\ref{tab:accuracy_comparison}).
These results indicate that current VLMs fall short on labeling ASMR videos as real rather than fake.


\textbf{Impact of ASMR prompt suite.}
As shown in Fig.~\ref{fig:bench ablation}(a), we compare the quality of generated videos with and without our ASMR prompt suite, where the baseline setting uses only a minimal instruction to ``generate the video as realistically as possible.''
We find that the ASMR prompt suite consistently improves realism
scores across models, especially on GPT-4-turbo and Gemini-2.5-pro, where the realism score is the $\text{Fakeness}_{\mathcal{G}}$.

\textbf{Preference \textit{versus} judgement.}
Despite our default setting, where the model is prompted to directly classify videos as real or fake, we also test a preference setting in which the model is given two videos, one real and one fake.  
Fig.~\ref{fig:bench ablation}(b) shows that pair preference achieves higher
realism scores than direct judgment across models and prompt settings.
Moreover, as seen in Fig.~\ref{fig:bench ablation}(c), we test this on two settings: “Pair,” where the fake video is generated from the same ASMR videos, and “Pair Shuffle,” where it is generated from different ASMR videos, indicating the pair and pair-shuffle settings have little impact. We also find that explicit reasoning trace enhances the VLMs detection accuracy.

\vspace{-0.5em}
\section{Qualitative Examples}
\label{sec: qualitative results}
\vspace{-1em}
As seen in Fig.~\ref{fig:example}, we visualize representative VLMs' reasoning traces for fake ASMR video cases.
\textbf{Watermark influence}: Gemini-2.5-Pro detects the “Sora2 watermark” and identifies the video as fake. However, once the watermark is removed, the model classifies the video as real, indicting sota VLMs rely on watermarks as a detection shortcut, rather than truly distinguishing real and fake ASMR videos based on perceptual cues.
\textbf{Audio-visual immersion influence}: Fake ASMR videos generated by Sora2 are initially classified as real by Gemini-2.5-Flash; however, after including audio in the inputs, the VLM correctly detects them as fake, showing that while Sora2 can produce visually vivid and realistic videos, it fails to generate ASMR content with convincing audio–visual coherence from a human perceptual perspective.
\textbf{High-quality ASMR fake videos}: ASMR video generated by Veo3.1-Fast appears highly realistic in terms of hand motion, object interaction, and lighting consistency, causing GPT-5 and Gemini-2.5-Pro to mistakenly classify it as real.
Additional qualitative examples are provided in App.\ref{app: AI-Generated Videos Examples}\&\ref{app: visualization examples}.

\vspace{-0.5em}

\section{Conclusion}
\vspace{-0.5em}
In this work, we introduce the topic of ASMR as a novel testbed to challenge and advance the frontiers of video understanding and generation.
Unlike existing benchmarks, ASMR content demands a degree of fine-grained audio-visual perception and sensory immersion that is uniquely sensitive to realism failures.
To facilitate research in this direction, we construct \ours, a comprehensive 1,500 high-quality ASMR videos curated from social media, paired with a specialized image and prompt suite for conditional video generation.
Our experiments demonstrate that even top-tier VLMs remain unreliable at detecting AI-generated ASMR videos, while current VGMs can produce ASMR videos that fool VLMs yet fail to deceive human annotators.
Our further observations and findings about the VLMs and VGMs in the ASMR domain provide valuable insights into the limitations of existing models and highlight promising future research directions.

\bibliographystyle{unsrt}
\bibliography{main}

\appendix

\input{appendix}

\end{document}

%% file: table/compare.tex
\begin{table*}[!t]
    \centering
    \caption{
    \textbf{Comparison of \ours and existing benchmarks.}
    }
    \resizebox{\linewidth}{!}{
    \begin{tabular}{lccccccccccc}
    \hline
   \multirow{2}{*}{Category}  & \multirow{2}{*}{Benchmark}  & \multicolumn{3}{c}{\textbf{Video Source}} & \multicolumn{3}{c}{\textbf{Participants}} & \multirow{2}{*}{\textbf{Eval. Prototype}} & \multicolumn{3}{c}{\textbf{Multi-Modal Inputs}} \\
      &  &  Real & Fake & Pair & VLM & VGM & Human &  & Text  & Image & Audio   \\
      \hline

\multirow{1}{*}{Video} 
& VBench~\cite{huang2024vbench}         & \errormark & \errormark & \errormark & \errormark & \correctmark & \correctmark & Acc. & \errormark & \errormark & \errormark \\
\multirow{2}{*}{Generation} & VideoPhy~\cite{bansal2024videophy}   & \errormark & \errormark  & \errormark & \errormark & \correctmark & \correctmark & Acc. & \errormark  & \errormark & \errormark \\
& PhyGenBench~\cite{phygenbench}       & \errormark & \errormark & \errormark & \errormark & \correctmark  & \correctmark & VLM-as-Judge & \errormark  & \errormark & \errormark \\
\hline
\multirow{1}{*}{Video} 
& SEED-Bench~\cite{seed_bench}        & \correctmark & \errormark & \errormark & \correctmark & \errormark &\correctmark & MCQ, Acc. & \correctmark   & \errormark & \errormark\\
\multirow{2}{*}{Understanding} & MV-Bench~\cite{li2024mvbench}       & \correctmark & \errormark & \errormark & \correctmark & \errormark &\errormark & MCQ, Acc. & \correctmark  & \errormark & \errormark\\
& TempCompass~\cite{liu2024tempcompass} & \correctmark  & \errormark & \errormark & \correctmark &\errormark & \correctmark & VLM-as-Judge, Acc. & \correctmark   & \errormark & \errormark\\
\hline
\multirow{2}{*}{AIGC} 
& GenVideo~\cite{gen_video}            & \correctmark & \correctmark & \errormark & \correctmark & \errormark &\errormark & Acc. & \errormark  & \errormark  & \errormark\\
\multirow{2}{*}{Detection} & LOKI~\cite{ye2024loki}              & \correctmark & \correctmark & \correctmark &\correctmark & \errormark & \correctmark & VLM-as-Judge, MCQ, Acc. & \correctmark & \errormark  & \errormark \\
& IPV-Bench~\cite{bai2025impossible}  & \correctmark & \correctmark & \errormark & \correctmark & \correctmark &\correctmark & VLM-as-Judge, MCQ, Acc. & \correctmark   & \errormark & \errormark \\
& {\textbf{\ours (Ours)}}  & \correctmark & \correctmark & \correctmark & \correctmark & \correctmark & \correctmark & \textbf{Under.-Gen. Adversarial Eval} & \correctmark & \correctmark  & \correctmark \\
    \hline
    \end{tabular}
    }
    \vspace{-1.5em}
    \label{tab:bench comparasion}
\end{table*}

%% file: table/tabl1.tex
\begin{table*}[t]
    \centering
    \caption{
    \textbf{Evaluation results for video understanding models.}
    A higher score indicates better performance. 
    \colorbox{gold}{\textcolor{black}{Gold}}, \colorbox{silver}{\textcolor{black}{silver}}, and 
    \colorbox{bronze}{\textcolor{black}{bronze}} denote the best, second, and third performers
    based on average F1 score. 
    }
    \vspace{-0.5em}
\resizebox{\linewidth}{!}{
    \begin{tabular}{l cc cc cc cc | cc | c}
\toprule
\multirow{2}{*}{Model}
      & \multicolumn{2}{c}{Veo3.1-Fast} 
      & \multicolumn{2}{c}{Wan2.2-A14B} 
      & \multicolumn{2}{c}{Opensora-V2} 
      & \multicolumn{2}{c|}{HunyuanVideo} 
      & \multicolumn{2}{c|}{\textbf{Average}} 
      & \multirow{2}{*}{Rank}\\
\cmidrule(lr){2-3}\cmidrule(lr){4-5}\cmidrule(lr){6-7}\cmidrule(lr){8-9}\cmidrule(lr){10-11}
      
      & Acc & F1 
      & Acc & F1 
      & Acc & F1 
      & Acc & F1 
      & Acc\,($\uparrow$) & F1\,($\uparrow$) 
      & \\
\midrule
      Random 
      & 50.00 & 50.00 & 50.00 & 50.00 & 50.00 & 50.00 & 50.00 & 50.00 
      & 50.00 & 50.00 & --\\
      Human  
      & 81.25 & 64.00 & 86.25 & 69.23 & 91.25 & 74.07 & 91.25 & 74.07 
      & 87.50 & \textbf{70.34} & \cellcolor{gold}{1}\\
\midrule
      \rowcolor{mygray}\multicolumn{12}{c}{\emph{Open-source Video Understanding Models}} \\
\midrule
      Qwen3-VL-8B          
      & 57.79 & 46.54 & 55.56 & 40.00 & 51.50 & 34.90 & 54.50 & 41.29 
      & 54.84 & 40.68 & 5\\
      Qwen3-VL-30B-A3B     
      & 51.08 & 31.01 & 49.44 & 23.53 & 47.09 & 21.88 & 49.74 & 28.57 
      & 49.34 & 26.25 & 7\\
      Qwen3-VL-235B-A22B   
      & 56.53 & 31.15 & 53.89 & 17.82 & 50.79 & 14.68 & 48.19 &  9.09 
      & 52.35 & 18.19 & 11\\
      GLM-4.5V             
      & 54.64 & 35.71 & 54.90 & 35.51 & 66.24 & 59.54 & 61.01 & 50.79 
      & 59.20 & \textbf{45.39} & \cellcolor{silver}{2}\\
\midrule
      \rowcolor{mygray}\multicolumn{12}{c}{\emph{Proprietary Video Understanding Models}} \\
\midrule
      GPT-4o-mini          
      & 52.50 & 27.64 & 53.68 & 15.09 & 53.00 & 20.34 & 50.50 & 12.39 
      & 52.42 & 18.87 & 10\\
      GPT-4o               
      & 51.50 & 19.33 & 55.26 & 22.60 & 56.50 & 17.86 & 56.50 & 25.53 
      & 54.94 & 21.33 & 9\\
      GPT-5                
      & 54.55 & 19.30 & 55.26 & 28.56 & 56.78 & 27.30 & 56.50 & 25.53 
      & 55.77 & 25.17 & 8\\
      Gemini-2.5-Flash     
      & 52.55 &  8.00 & 53.55 & 33.59 & 55.15 & 38.40 & 53.06 & 36.06 
      & 53.58 & 29.01 & 6\\
      Gemini-2.5-Pro       
      & 56.00 & 25.60 & 59.09 & 37.93 & 62.30 & 48.12 & 65.76 & 55.32 
      & 60.79 & 41.74 & 4\\
      Gemini-3-Pro-Preview 
      & 77.89 & 25.60 & 57.67 & 46.48 & 65.83 & 48.12 & 80.90 & 55.32 
      & 70.57 & \textbf{43.88} & \cellcolor{bronze}{3}\\
\bottomrule
    \end{tabular}
}
    \vspace{-1.5em}
    \label{tab:table2}
\end{table*}



%% file: table/opensora_ablation.tex
\begin{table}[!t]
    \centering
    \captionsetup{font=small}
    \caption{\textbf{Evaluation results for video generation models under different generation settings.}
    Image means whether to use our ASMR image suite.
    A lower score indicates better performance. 
    \colorbox{gold}{\textcolor{black}{Gold}}, \colorbox{silver}{\textcolor{black}{silver}}, and 
    \colorbox{bronze}{\textcolor{black}{bronze}} denote the best, second, and third performers.
    }
    \resizebox{\linewidth}{!}{
    \begin{tabular}{l|c|cccc|cc}
    \toprule
    Generation Type & ASMR Image Suite & GPT-4o-mini & GPT-4o & Gemini-2.5-Flash & Gemini-2.5-Pro & \textbf{Avg.\,($\downarrow$)} & Rank\\
    \hline
          \rowcolor{mygray}\multicolumn{8}{c}{\emph{Opensora-V2}~\cite{peng2025open}} \\
    \hline
    Text2Vid        & \errormark  & 14.00 & 10.00 & 28.72 & 39.18 & 22.98 & 8 \\
    ImgText2Vid     & \correctmark  & 12.00 & 15.00 & 30.21 & 35.16 & 23.59 & 9 \\
    Text2Img2Vid    & \errormark  & 14.00 & 18.00 & 45.36 & 43.75 & 30.28 & 10 \\
    \hline
    \rowcolor{mygray}\multicolumn{8}{c}{\emph{Wan2.2}~\cite{wan2025wan}} \\
    \hline
    Text2Vid-14B    & \errormark  & 12.00 & 13.00 & 24.47 & 21.74 & 17.80 & 5 \\
    ImgText2Vid-14B & \correctmark & 8.89 & 7.78 & 23.53 & 26.19 & 16.10 & \cellcolor{bronze}{3} \\
    ImgText2Vid-5B  & \correctmark & 7.00 & 13.00 & 30.53 & 33.33 & 20.97 & 6 \\
    \hline
    \rowcolor{mygray}\multicolumn{8}{c}{\emph{HunyuanVideo}~\cite{kong2024hunyuanvideo}} \\
    \hline
    Text2Vid        & \errormark  & 8.00 & 7.00 & 25.51 & 18.56 & 14.77 & \cellcolor{silver}{2} \\
    ImgText2Vid     & \correctmark & 7.00 & 15.00 & 26.53 & 42.39 & 22.73 & 7 \\
    \hline
    \rowcolor{mygray}\multicolumn{8}{c}{\emph{Sora2}~\cite{sora2}} \\
    \hline
    Text2Vid        & \errormark  & 16.00 & 9.00  & 100.00 & 97.89 & 55.72 & 12 \\
    ImgText2Vid     & \correctmark & 8.25  & 3.09  & 95.79  & 79.17 & 46.58 & 11 \\
    \rowcolor{lightpurple}
    + Audio         & \correctmark & 8.25  & 3.09  & 97.89  & 88.66 & 49.47\textsubscript{$\textcolor{my_red}{\scriptsize +2.89}$} & 11 \\
    - Watermark     & \correctmark & 8.25  & 6.19  & 25.00  & 24.74 & 16.55 & 4 \\
    \hline
    \rowcolor{mygray}\multicolumn{8}{c}{\emph{StepVideo}~\cite{ma2025step}} \\
    \hline
    Text2Vid        & \errormark  & 84.00 & 92.00 & 73.68 & 86.81 & 83.62 & 13 \\
    \hline
    \rowcolor{mygray}\multicolumn{8}{c}{\emph{Veo-3.1-fast}~\cite{veo3}} \\
    \hline
    ImgText2Vid     & \correctmark & 11.00 & 5.00 & 16.16 & 17.00 & 12.54 & \cellcolor{gold}{1} \\
    \rowcolor{lightpurple}
    + Audio         & \correctmark & 11.00 & 5.00 & 19.20 & 25.00 & 15.05\textsubscript{$\textcolor{my_red}{\scriptsize +2.51}$} & \cellcolor{gold}{1} \\
    \bottomrule
    \end{tabular}
    }
    \label{tab:v_gen}
    \vspace{-1.5em}
\end{table}

%% file: appendix.tex
\newpage
\begin{center}
	\LARGE \bf {Appendix}
\end{center}

\etocdepthtag.toc{mtappendix}
\etocsettagdepth{mtchapter}{none}
\etocsettagdepth{mtappendix}{subsubsection}
\tableofcontents
\newpage

\section{AI-Generated Videos Examples}
\label{app: AI-Generated Videos Examples}

\subsection{Easy Split Examples for Sora2}

\begin{figure*}[!t]
    \centering
    
    \begin{subfigure}{\linewidth}
        \centering
        \includegraphics[width=\linewidth]{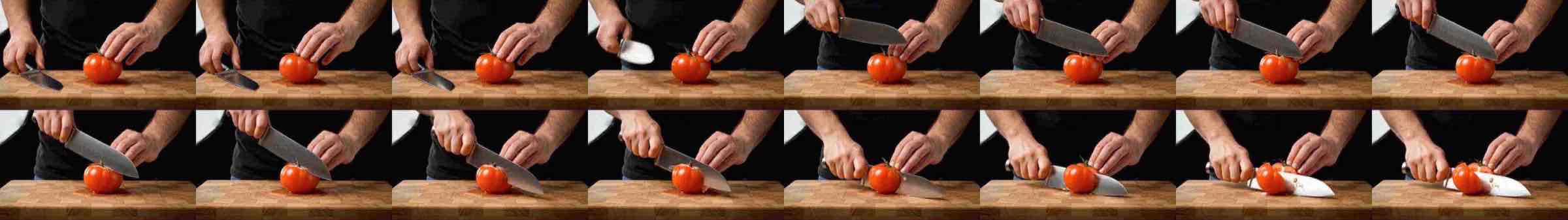}
        \caption{Real ASMR video for cutting tomatoes.}
        \label{fig:top}
    \end{subfigure}

    \begin{subfigure}{\linewidth}
        \centering
        \includegraphics[width=\linewidth]{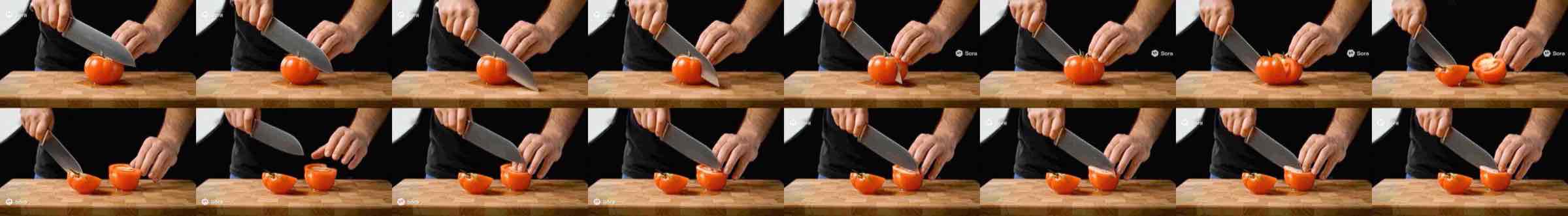}
        \caption{Sora2 generated video for cutting tomatoes.}
        \label{fig:bottom}
    \end{subfigure}

    \begin{subfigure}{\linewidth}
        \centering
        \includegraphics[width=\linewidth]{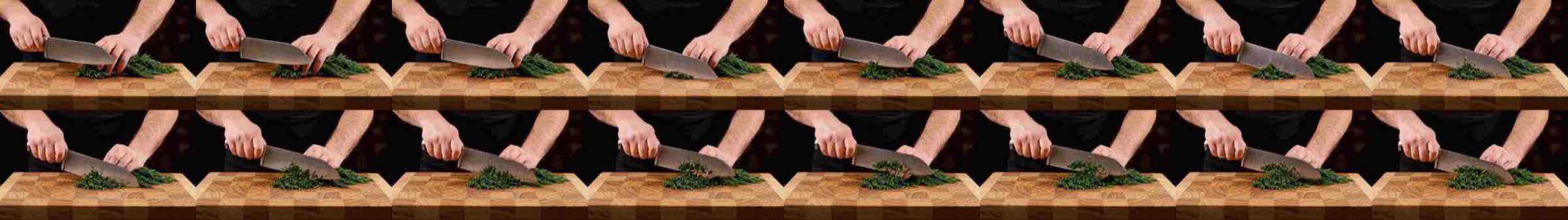}
        \caption{Real ASMR video for cutting vegetables.}
        \label{fig:top}
    \end{subfigure}




    \begin{subfigure}{\linewidth}
        \centering
        \includegraphics[width=\linewidth]{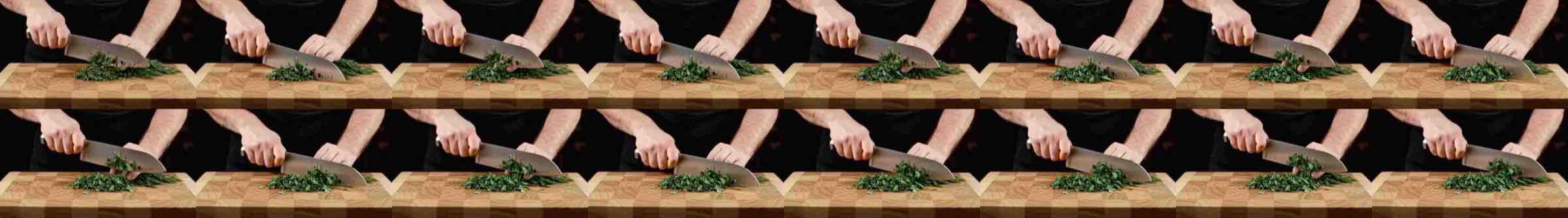}
        \caption{Sora2 generated video for cutting vegetables.}
        \label{fig:bottom}
    \end{subfigure}

    \begin{subfigure}{\linewidth}
        \centering
        \includegraphics[width=\linewidth]{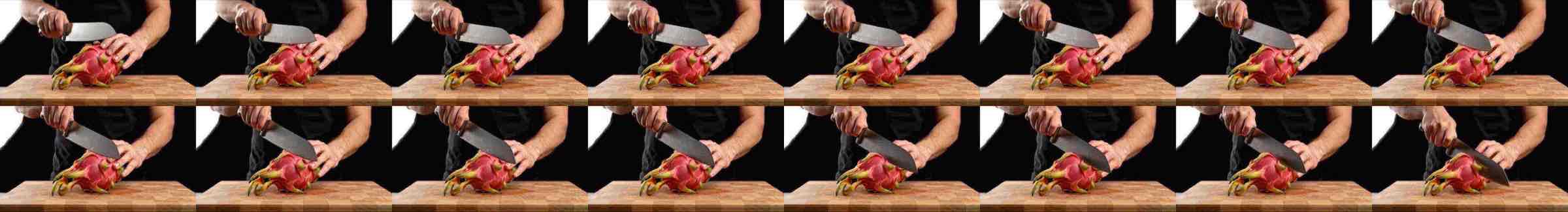}
        \caption{Real ASMR video for cutting pitaya.}
        \label{fig:top}
    \end{subfigure}


    \begin{subfigure}{\linewidth}
        \centering
        \includegraphics[width=\linewidth]{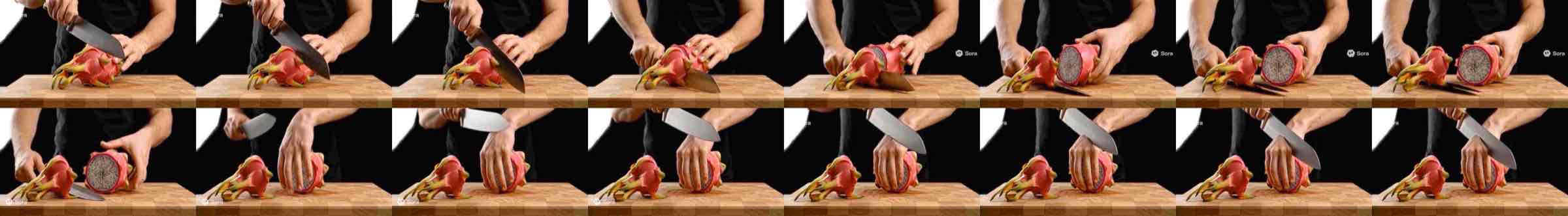}
        \caption{Sora2 generated video for cutting pitaya.}
        \label{fig:middle}
    \end{subfigure}

    \caption{Real v.s. Fake for the easy split of \ours. The sora2-generated videos are conditioned on our ASMR prompt suite and ASMR image suite.}
    \label{fig: real vs fake 3}
\end{figure*}

Figure~\ref {fig: real vs fake 3} illustrates representative examples from the easy split of \ours, pairing real ASMR videos with Sora2-generated counterparts across three food-cutting scenarios: tomatoes, vegetables, and pitaya. 
The Sora2-generated videos in this split exhibit conspicuous physical artifacts that betray their synthetic origin. 
In the tomato-cutting sequence, the tomato retains its original shape even after being sliced — a clear violation of basic physical plausibility. 
More strikingly, the pitaya-cutting sequence contains a shot where the knife visibly cuts into the human hand, indicating a severe failure in modeling real-world object interactions. 
While such artifacts are immediately apparent to human annotators, video understanding models surprisingly struggle to detect them reliably, suggesting that current VLMs lack the fine-grained physical commonsense required to identify even obvious generation failures.

\subsection{Hard Split Examples for Sora2 and Veo3.1}

\begin{figure*}[!t]
    \centering
    
    \begin{subfigure}{\linewidth}
        \centering
        \includegraphics[width=\linewidth]{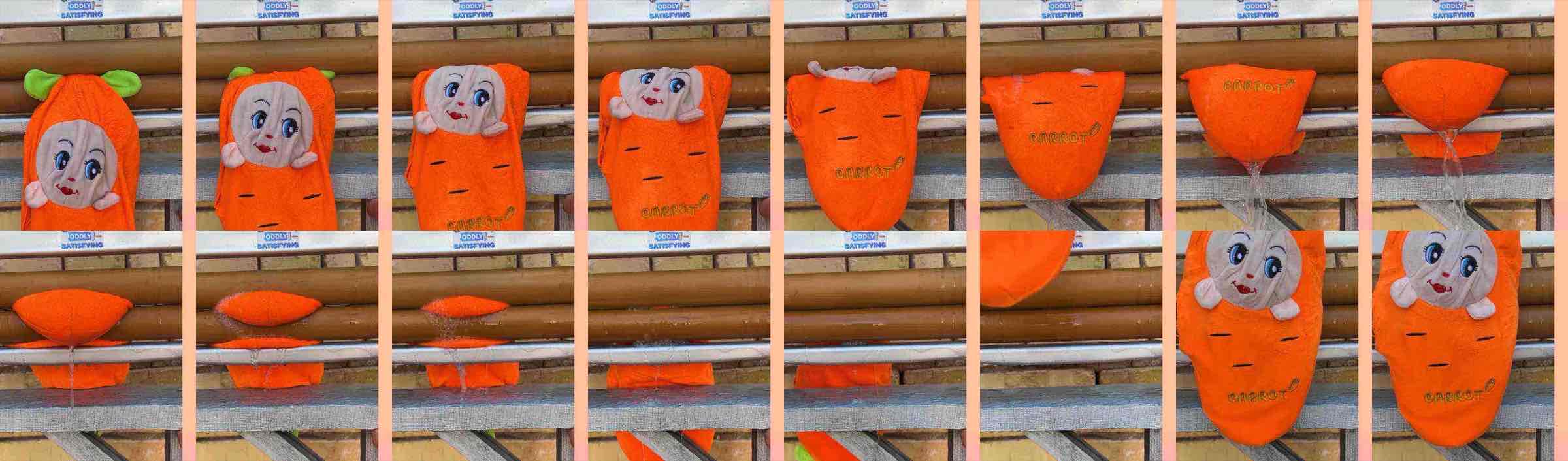}
        \caption{Real ASMR video.}
        \label{fig:top}
    \end{subfigure}

    \vspace{0.4em}

    \begin{subfigure}{\linewidth}
        \centering
        \includegraphics[width=\linewidth]{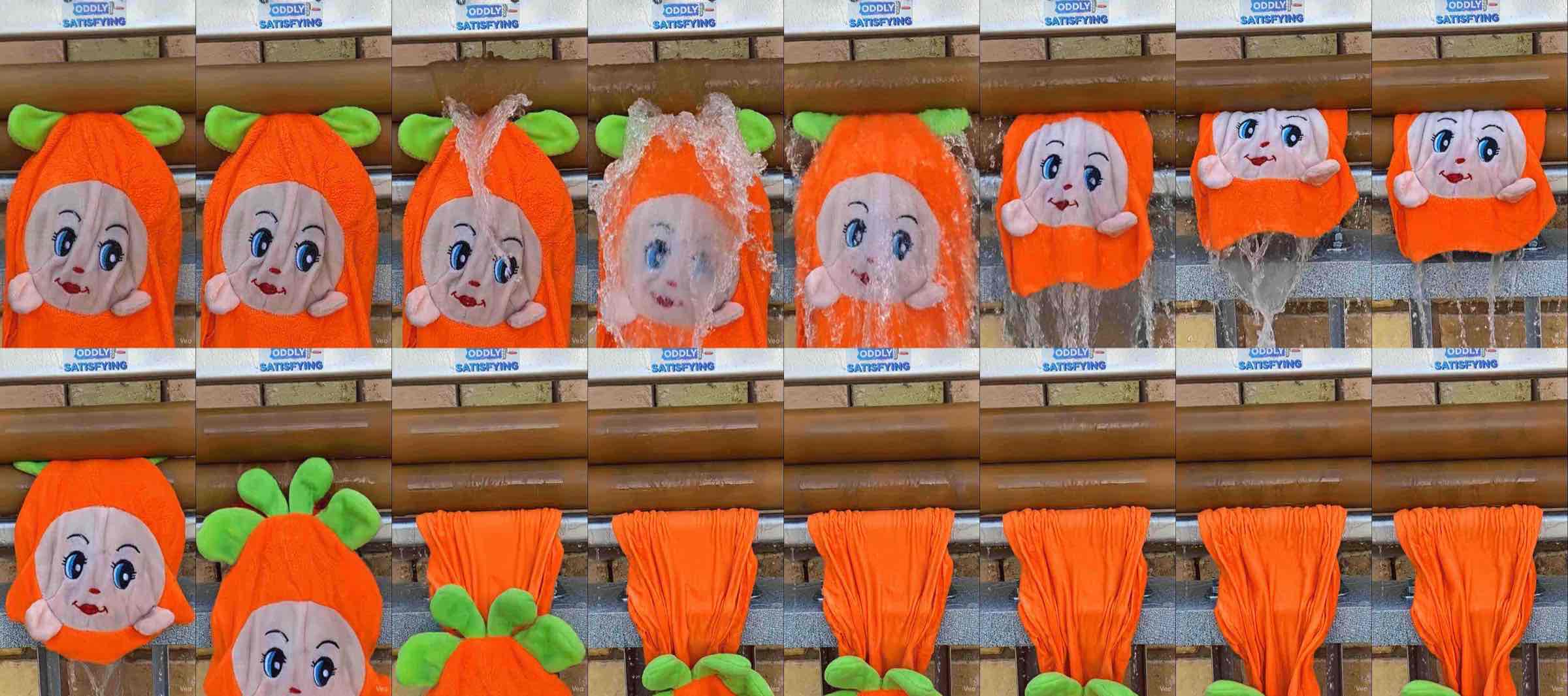}
        \caption{Veo3.1-fast generated video.}
        \label{fig:middle}
    \end{subfigure}

     \vspace{0.4em}

    \begin{subfigure}{\linewidth}
        \centering
        \includegraphics[width=\linewidth]{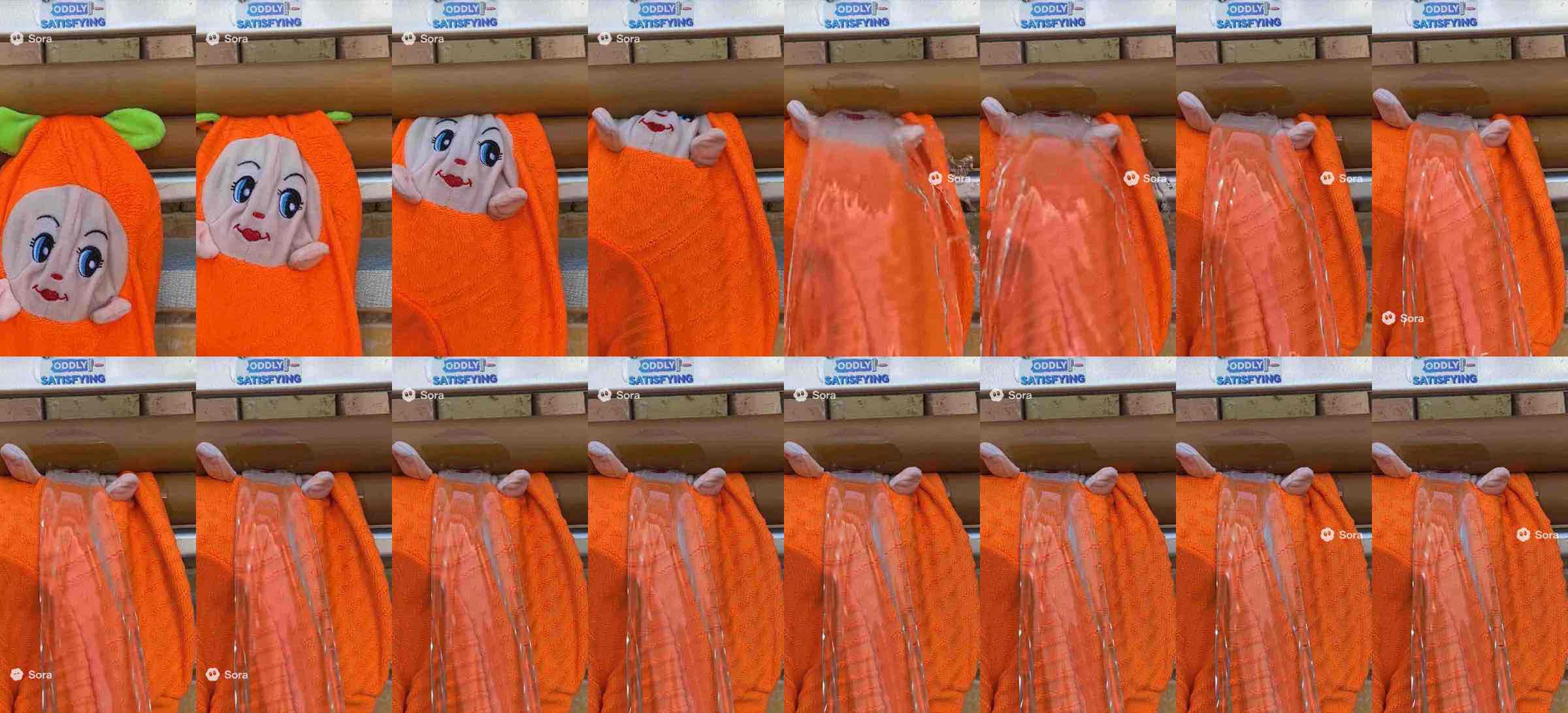}
        \caption{Sora2 generated video.}
        \label{fig:bottom}
    \end{subfigure}

    \caption{Real v.s. Fake on \ours of the hard split with our ASMR prompt suite and ASMR image suite.}
    \label{fig: real vs fake 1}
\end{figure*}

\begin{figure*}[!t]
    \centering
    
    \begin{subfigure}{\linewidth}
        \centering
        \includegraphics[width=\linewidth]{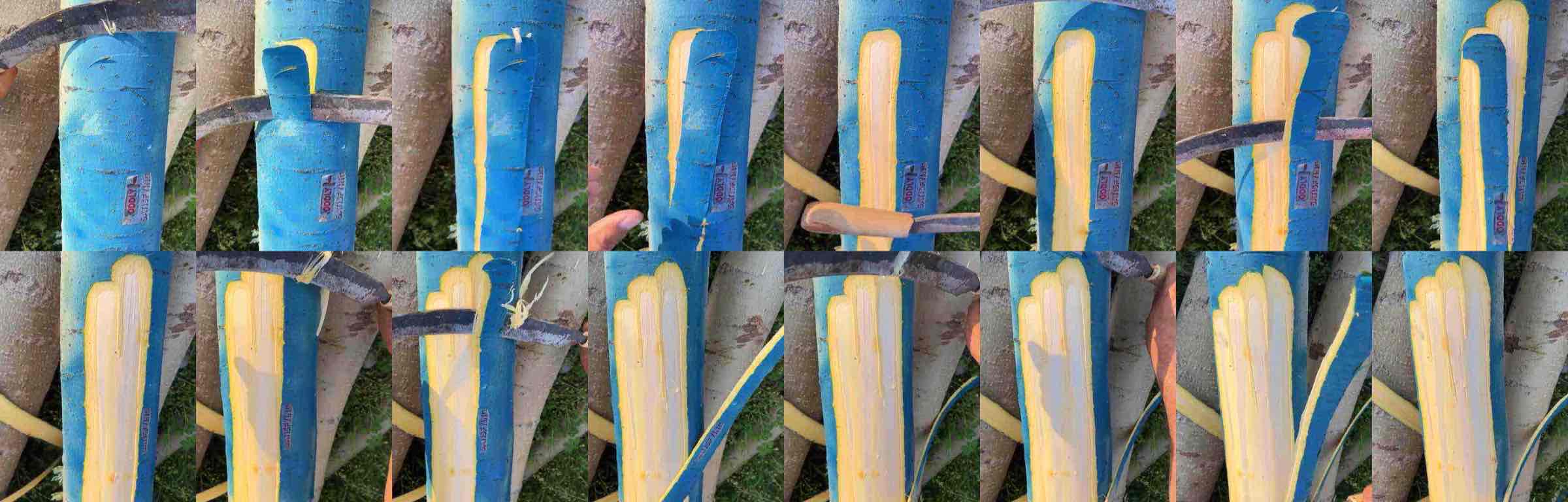}
        \caption{Real ASMR video.}
        \label{fig:top}
    \end{subfigure}


    \begin{subfigure}{\linewidth}
        \centering
        \includegraphics[width=\linewidth]{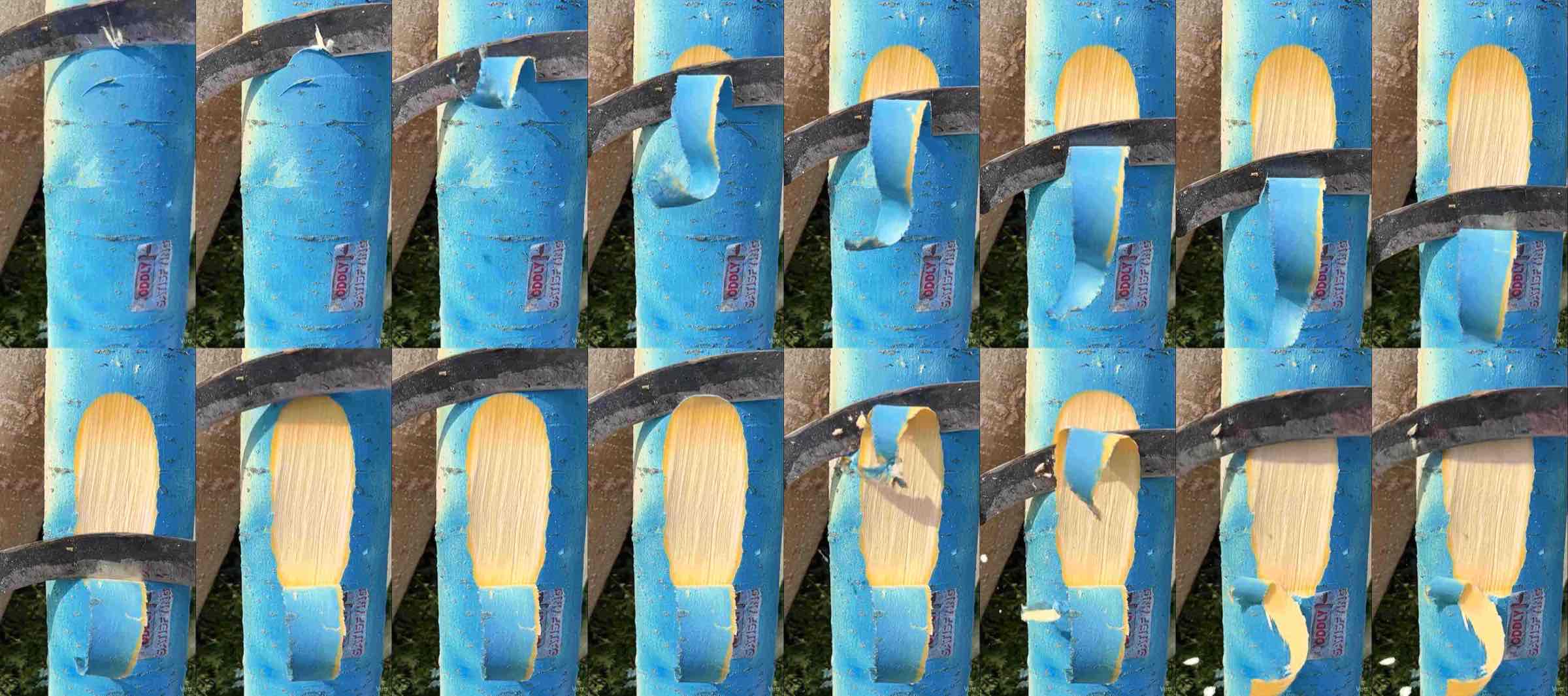}
        \caption{Veo3.1-fast generated video.}
        \label{fig:middle}
    \end{subfigure}


    \begin{subfigure}{\linewidth}
        \centering
        \includegraphics[width=\linewidth]{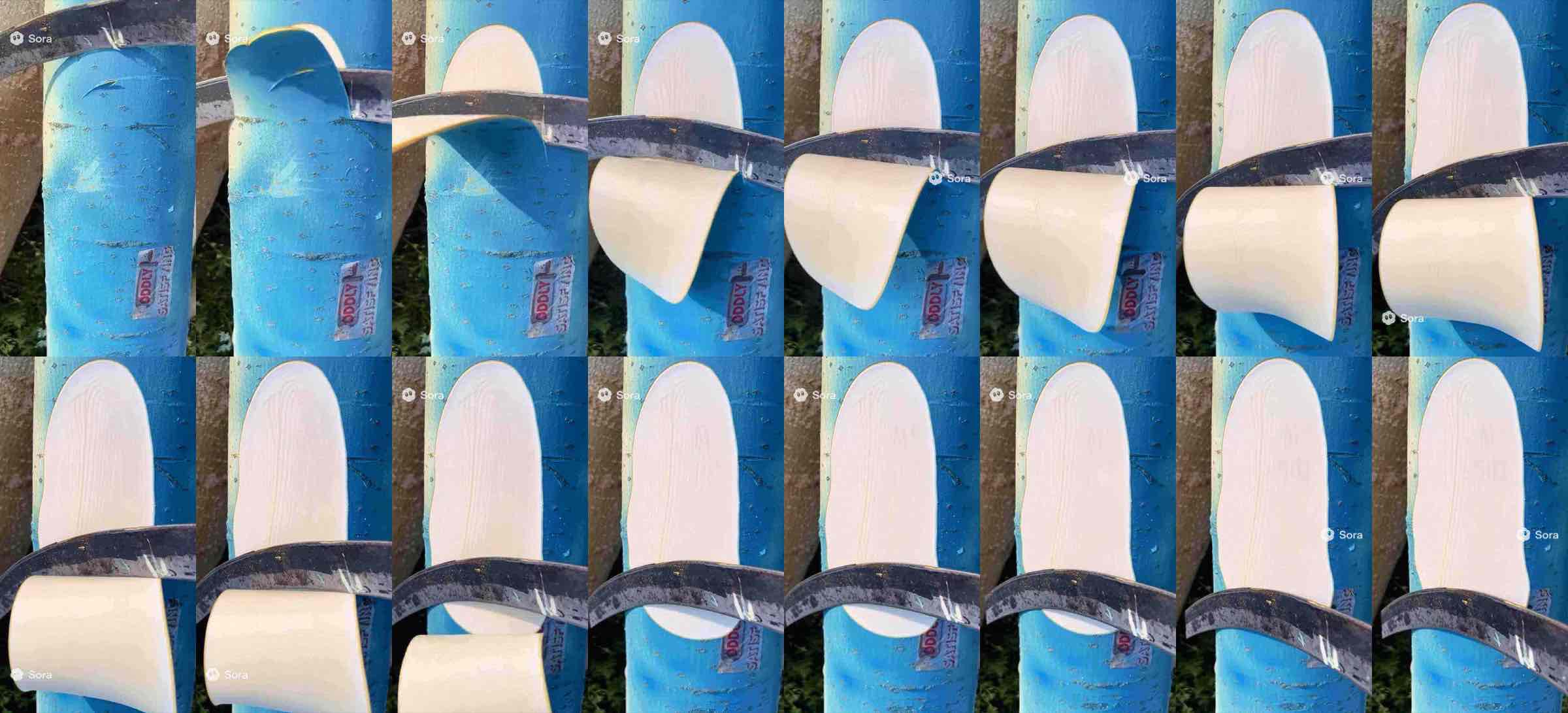}
        \caption{Sora2 generated video.}
        \label{fig:bottom}
    \end{subfigure}

    \caption{Real v.s. Fake on \ours of the hard split with our ASMR prompt suite and ASMR image suite.}
    \label{fig: real vs fake 2}
\end{figure*}

\begin{figure*}[!t]
    \centering
    
    \begin{subfigure}{\linewidth}
        \centering
        \includegraphics[width=\linewidth]{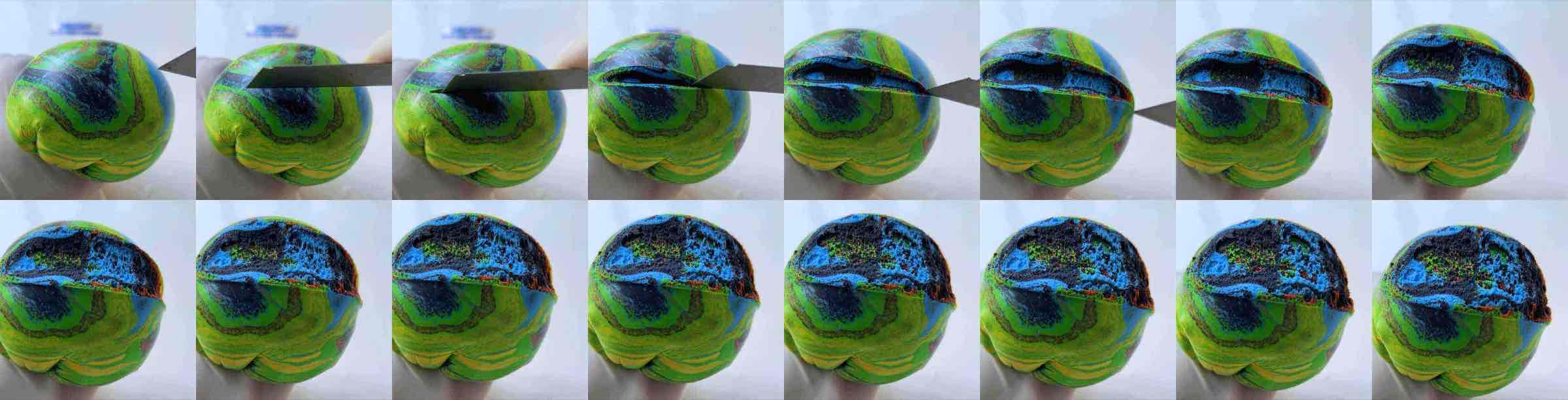}
        \caption{Real ASMR video.}
        \label{fig:top}
    \end{subfigure}


    \begin{subfigure}{\linewidth}
        \centering
        \includegraphics[width=\linewidth]{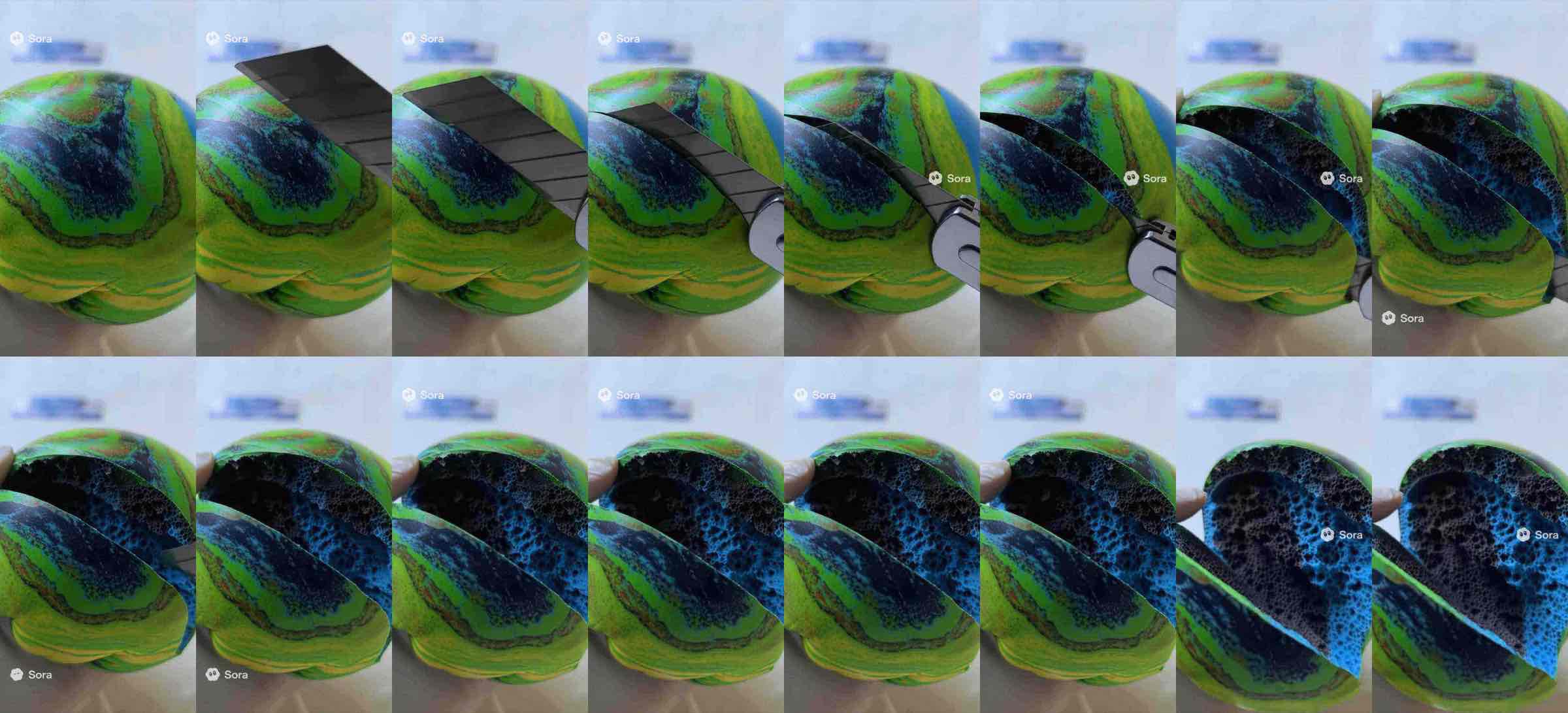}
        \caption{Sora2 generated video.}
        \label{fig:middle}
    \end{subfigure}
    \caption{Real v.s. Fake on \ours of the hard split with our ASMR prompt suite and ASMR image suite.}
    \label{fig: real vs fake 6}
\end{figure*}

\begin{figure*}[!h]
    \centering
    
    \begin{subfigure}{\linewidth}
        \centering
        \includegraphics[width=\linewidth]{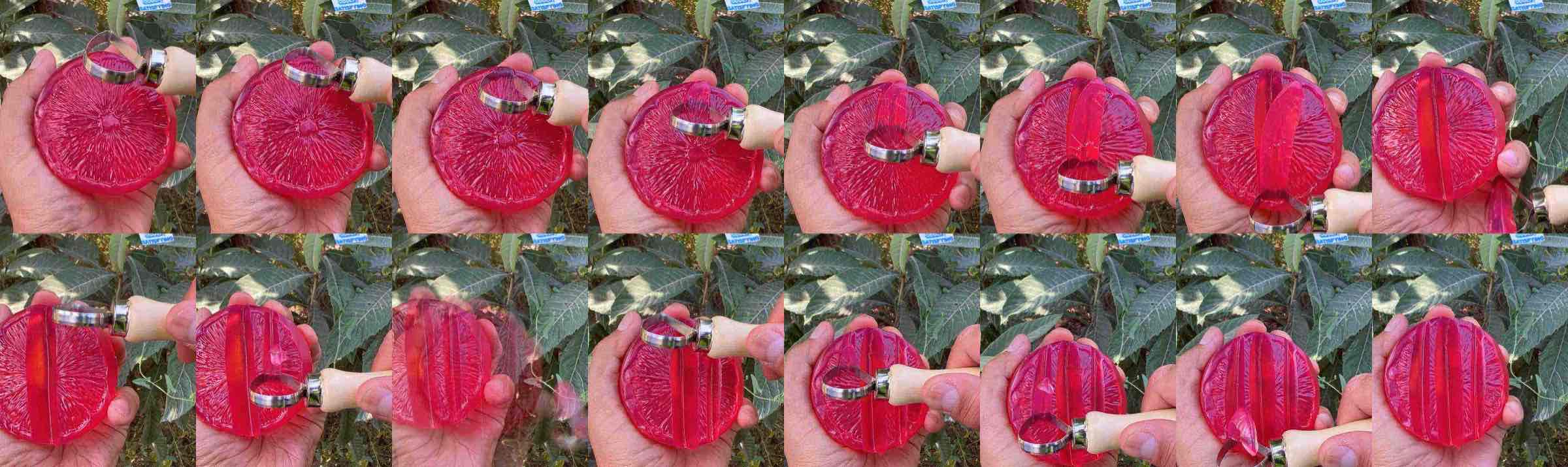}
        \caption{Real ASMR video.}
        \label{fig:top}
    \end{subfigure}


    \begin{subfigure}{\linewidth}
        \centering
        \includegraphics[width=\linewidth]{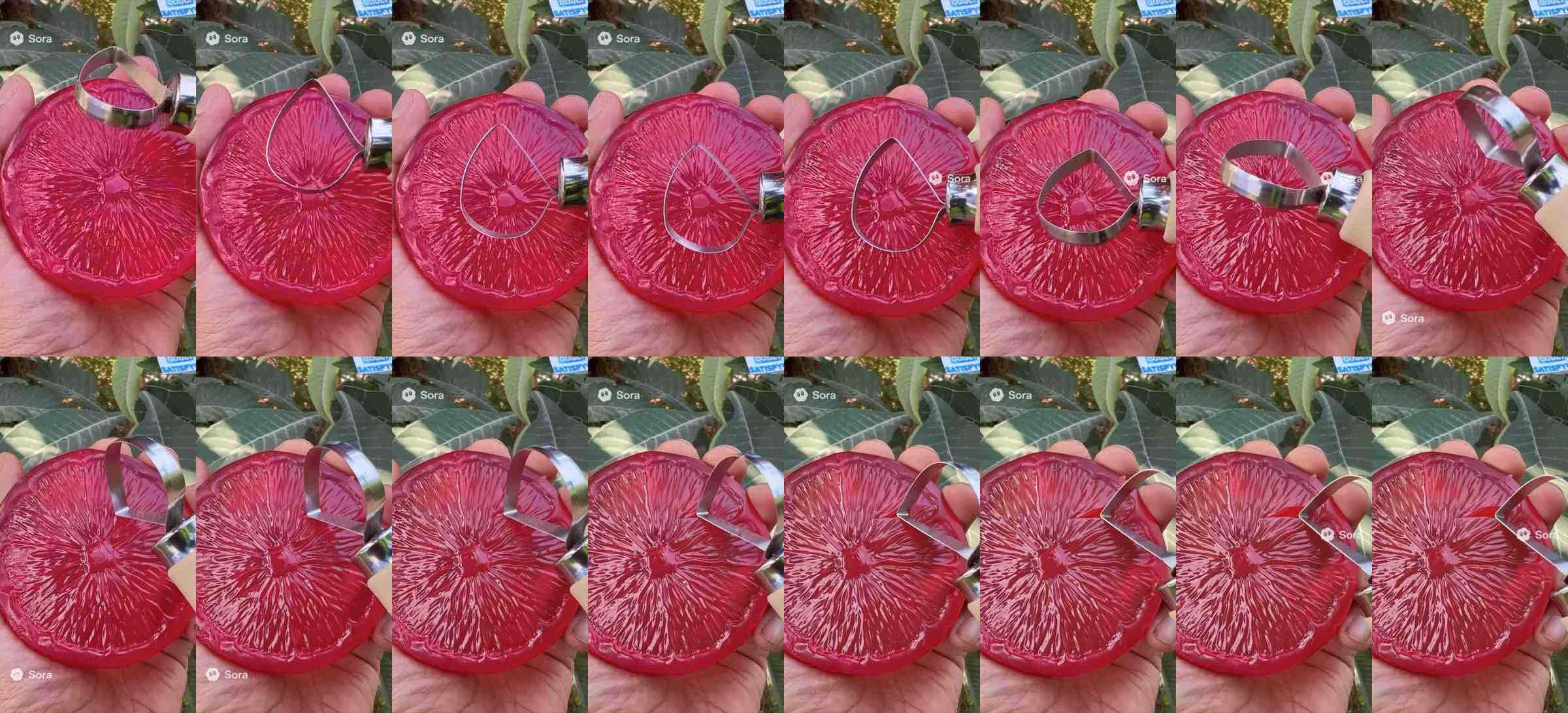}
        \caption{Sora2 generated video.}
        \label{fig:middle}
    \end{subfigure}
    \caption{Real v.s. Fake on \ours of the hard split with our ASMR prompt suite and ASMR image suite.}
    \label{fig: real vs fake 7}
\end{figure*}

    






Compared to the easy split, the hard split encompasses a broader and more diverse range of ASMR scenarios beyond simple food-cutting, including squeezing water out of toys, chopping branches, and slicing bath foam balls. These scenes demand richer visual and auditory coherence — for instance, the subtle sound of water being squeezed, or the soft crackling of dissolving foam — making authenticity detection substantially more challenging.

As illustrated in Figures~\ref{fig: real vs fake 1}--\ref{fig: real vs fake 7}, Veo3.1-Fast generally produces videos that closely follow the ASMR prompt suite and image suite, yielding visually plausible results that are difficult to distinguish from real footage. In contrast, Sora2-generated videos frequently deviate from the provided instructions, failing to faithfully reproduce the intended ASMR scenarios. More notably, the audio in Sora2 outputs is dominated by human voices rather than the natural object-interaction sounds expected in ASMR content, suggesting that Sora2 may be overfitting to human-centric audio patterns present in its training data.

\section{Reasoning Traces Examples.}
\label{app: visualization examples}

Below we show the ASMR video frames from \ours and the reasoning process along with the answers for qualitative analysis.

\subsection{Comparison with and without Audio for Veo3.1}

\begin{figure*}[!t]
    \centering
    \includegraphics[width=\linewidth]{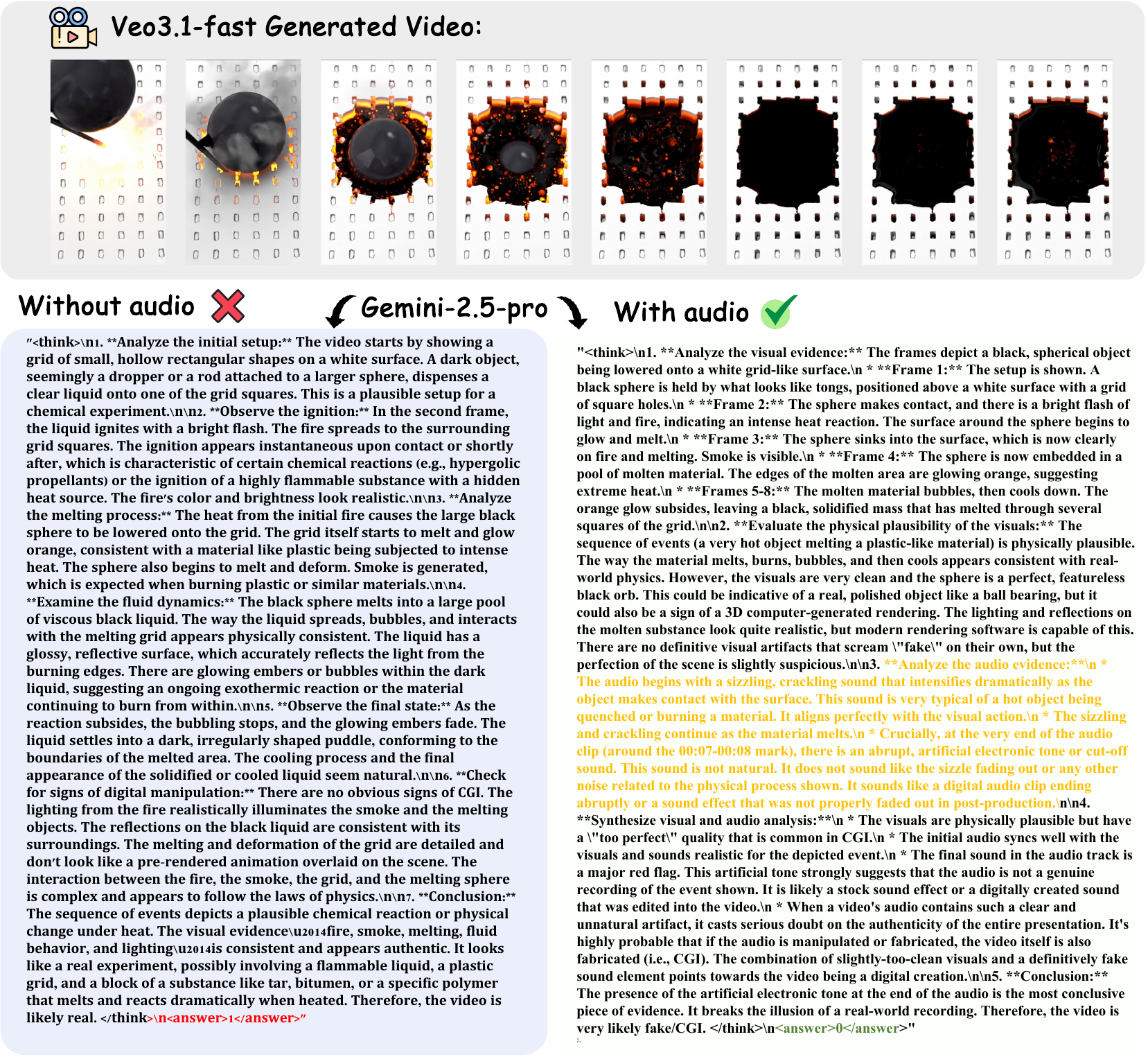}
    \caption{Gemini-2.5-pro on veo3.1-fast generated videos, with and without audio. After adding the audio, the VLM detects the video to be fake.
    Red marks the wrong answer.
    Green marks the right answer.
    Yellow marks the audio-related thinking process.}
    \label{fig: audio 1}
\end{figure*}

Figure~\ref{fig: audio 1}: Veo3.1-fast generated videos with and without audio evaluation, where with audio is detected to be generated, while without audio is real.

\subsection{Comparison with and without Audio for Sora2}

\begin{figure*}[!t]
    \centering
    \includegraphics[width=\linewidth]{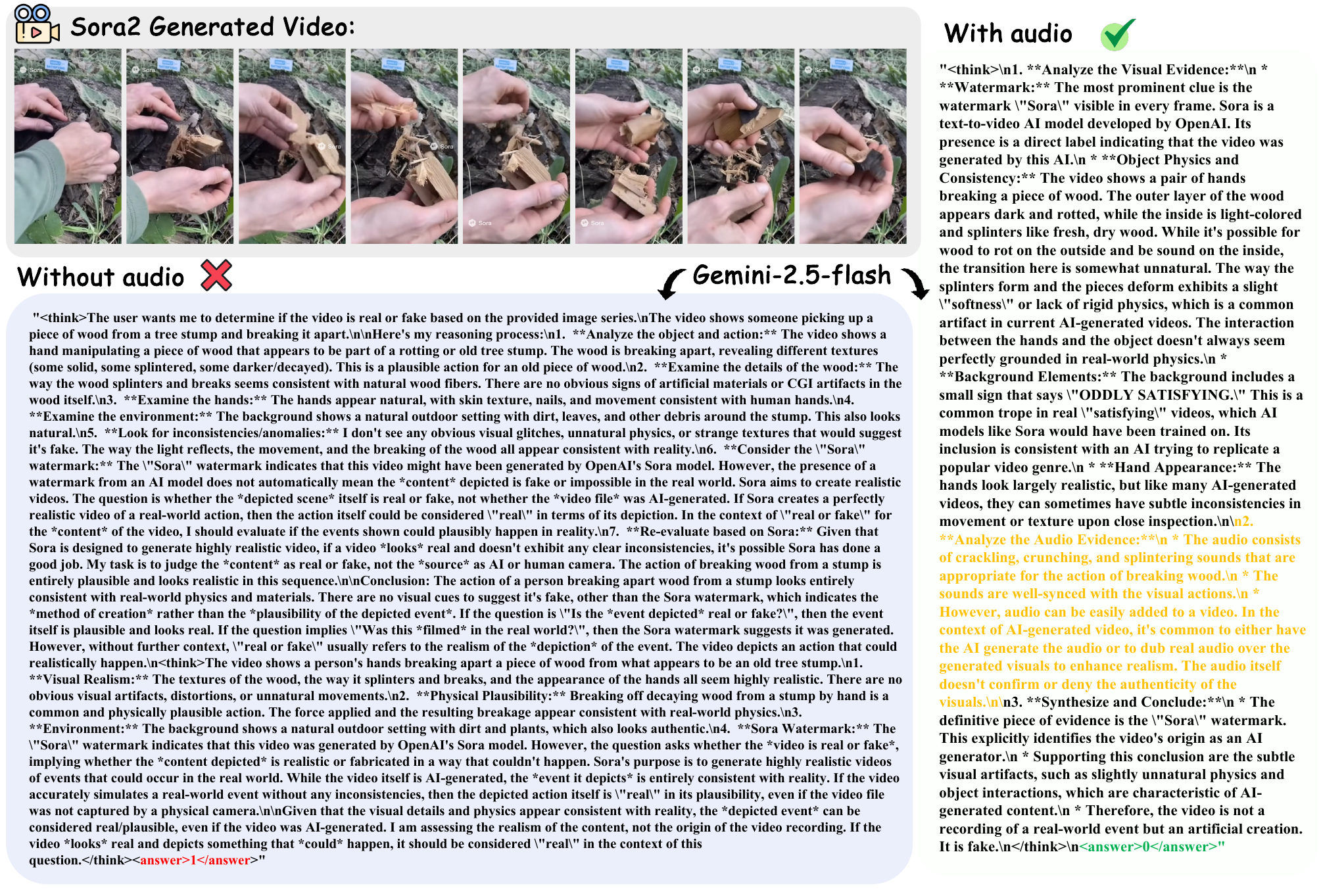}
    \caption{Gemini-2.5-flash on sora2 generated videos, with and without audio. After adding the audio, the VLM detects the video to be fake.
    Red marks the wrong answer.
    Green marks the right answer.
    Yellow marks the audio-related thinking process.}
    \label{fig: audio 2}
\end{figure*}

Figure~\ref{fig: audio 2}: Sora2 generated videos with and without audio evaluation, where with audio is detected to be generated, while without audio is real.

\section{Broader Impacts}
\label{sec: broader impacts}

This paper presents work whose goal is to advance the field of Machine Learning. There are many potential societal consequences of our work, none of which we feel must be specifically highlighted here.

\section{Limitation}
\label{sec: limitation}

First, while \ours centers on ASMR as its evaluation domain, we emphasize that ASMR should be understood as a rigorous scientific testbed rather than a niche sub-domain. As foundation models rapidly advance, assessing fine-grained physical fidelity and precise audio-visual synchronization has become an inevitable next frontier, and we are the first to pioneer this space in generative AI, utilizing ASMR as a unique "microscope" to expose subtle hallucinated artifacts in current models. 
Second, beyond standard judgment tasks, \ours further incorporates preference-based evaluation and in-depth CoT analysis, showing that models explicitly reason about audio-visual incoherence rather than making blind classifications, which clearly sets our protocol apart from conventional classification tasks. 
Moreover, while human annotation may introduce subjective bias, our evaluation framework leverages a diverse set of state-of-the-art VLMs and VGMs in a mutually supervisory manner, leading to a more rigorous and reliable assessment. 
Finally, this work is fundamentally proposed as a benchmark and evaluation protocol rather than a methodological advance; in the generative AI era, where evaluating fine-grained realism remains a critical bottleneck, providing a robust, scalable dataset and an automated evaluation paradigm is an essential contribution for the community's future modeling efforts.

\section{Evaluation metric}
\label{app: evaluation metric}


We formulate the adversarial understanding-generation evaluation as a two-player game between a \emph{video generation} model \(\mathcal{G}\) and an \emph{video understanding} model \(\mathcal{U}\).

Given a real ASMR video \(\mathcal{V}_{\text{real}}\), the video generation model $\mathcal{G}$ produces a fake video
\[
\mathcal{V}^g_{\text{fake}} = \mathcal{G}(\mathcal{I}, \mathcal{T}),
\]
conditioned on a reference image \(\mathcal{I}\) from our ASMR image suite and a text prompt \(\mathcal{T}\) from our ASMR prompt suite. 
The understanding model then predicts whether an input video \(\mathcal{V}\) is real or fake. 
The model is instructed to first perform explicit reasoning and then output the final answer.
All outputs are parsed by extracting the numeric label enclosed in
\texttt{<answer> number </answer>} using regular-expression matching where 0 is the fake video and 1 is the real; only valid predictions are considered.

\subsection{Understanding Model Detection Accuracy.}
For each video understanding model \(\mathcal{U}\), the model outputs a binary judgment
\[
\mathcal{U}(\mathcal{V}) \in \{0,1\},
\]
where \(1\) denotes a real video and \(0\) denotes a fake video. The model is evaluated on the full set
\[
\mathcal{V}^{g}_{\text{full}} = \{\mathcal{V}_{\text{real}}\} \cup \{\mathcal{V}^{g}_{\text{fake}}\},
\]
where real and fake videos are balanced in equal numbers to avoid numerical bias when computing classification accuracy. The detection accuracy is defined as
\[
\text{Acc}^{g}_{\mathcal{U}}
= \frac{1}{|\mathcal{V}^{g}_{\text{full}}|}
\sum_{\mathcal{V} \in \mathcal{V}^{g}_{\text{full}}}
\mathbf{1}\!\left[\mathcal{U}(\mathcal{V}) = y(\mathcal{V})\right],
\]
where \(y(\mathcal{V}) = 1\) for real videos and \(y(\mathcal{V}) = 0\) for fake videos.

For a comprehensive evaluation, we aggregate the detection performance of an understanding model across multiple video generation models \(\mathcal{G}\), formulated as
\[
\text{Acc}_{\mathcal{U}}
= \frac{1}{|\mathcal{G}|}
\sum_{g \in \mathcal{G}}
\text{Acc}^{g}_{\mathcal{U}} .
\]


\paragraph{F1 Score Computation}
We adopt F1 score alongside accuracy to better reflect model discriminability in the judgement task, where each model must classify a video as either \emph{real} or \emph{AI-generated} (fake).
We define \emph{AI-generated video} as the positive class. Given a set of real videos and videos produced by a video generation model, a video understanding model assigns a binary prediction to each sample.
For each video generation model $g$, we construct a confusion matrix as follows:

\begin{itemize}
    \item \textbf{TP} (True Positive): AI-generated videos correctly predicted as fake.
    \item \textbf{FP} (False Positive): Real videos incorrectly predicted as fake.
    \item \textbf{FN} (False Negative): AI-generated videos incorrectly predicted as real.
    \item \textbf{TN} (True Negative): Real videos correctly predicted as real.
\end{itemize}

Note that FP and TN are shared across all video generation models, as the real video set is fixed.
Precision, Recall, and F1 score are computed as:

\begin{equation}
    \text{Precision} = \frac{\text{TP}}{\text{TP} + \text{FP}}, \quad
    \text{Recall} = \frac{\text{TP}}{\text{TP} + \text{FN}}, \quad
    \text{F1} = \frac{2 \times \text{Precision} \times \text{Recall}}{\text{Precision} + \text{Recall}}
\end{equation}

For each video understanding model, we compute per-generator F1 scores across the $N$ video generation models evaluated, then report the macro-average:

\begin{equation}
    \overline{\text{F1}} = \frac{1}{N} \sum_{i=1}^{N} \text{F1}_i
\end{equation}

\paragraph{Why F1 over accuracy.}
Accuracy can be misleading in this setting: a model that predicts \emph{real} for all inputs achieves 50\% accuracy by construction (given a balanced test set), yet has zero discriminative power. F1 score is sensitive to both missed detections (low Recall) and false alarms (low Precision), making it a more informative measure of whether a model can genuinely distinguish AI-generated videos from real ones.

\subsection{Generation Model Realism Scores.}
For each video generation model \(\mathcal{G}\), the evaluation is conducted solely on its generated fake videos \(\mathcal{V}^{g}_{\text{fake}}\). The generation quality is measured by the rate at which these videos are detected as fake by a video understanding model \(\mathcal{U}\), defined as
\[
\text{Fakeness}^{u}_{\mathcal{G}}
= \frac{1}{|\mathcal{V}^{g}_{\text{fake}}|}
\sum_{\mathcal{V} \in \mathcal{V}^{g}_{\text{fake}}}
\mathbf{1}\!\left[\mathcal{U}(\mathcal{V}) = 0\right],
\]
where a lower value indicates that the generated videos are harder to distinguish from real ones.

For a comprehensive evaluation, we aggregate the detection results across multiple video understanding models \(\mathcal{U}\), yielding the overall detection rate for a generation model:
\[
\text{Fakeness}_{\mathcal{G}}
= \frac{1}{|\mathcal{U}|}
\sum_{u \in \mathcal{U}}
\text{Fakeness}^{u}_{\mathcal{G}} .
\]

Under this definition, a higher \(\text{Acc}_{\mathcal{U}}\) indicates a stronger video understanding model, which can better distinguish fake ASMR videos in \ours, whereas a lower \(\text{Fakeness}_{\mathcal{G}}\) reflects a more realistic video generation model, capable of producing ASMR videos that are more vivid and closer to those in \ours.
This unified formulation enables a dynamic and adversarial evaluation setting, in which generation and understanding models can iteratively improve through mutual interaction.

\section{Video generation settings}
\label{sec: video generation settings}
We adopt the default OpenSora V2 GitHub implementation to generate ASMR videos using our ASMR prompt suite and ASMR image suite. The generation parameters are summarized in table~\ref{tab:opensorav2}. 
We denote i2v\_head, t2v, and t2i2v as the image-to-video, text-to-video, and text-to-image-to-video generation settings, respectively.
\begin{table}[!t]
\centering
\small
\caption{OpensoraV2 Sampling configuration used for video generation. The i2v\_head, t2v, and t2i2v represent image2video, text2video and text2image2video generation settings.}
\begin{tabular}{ll}
\hline
\textbf{Parameter} & \textbf{Value} \\
\hline
batch\_size & 4 \\
dtype & bf16 \\
Resolution & 256px \\
Aspect ratio & 16:9 \\
Number of frames & 129 \\
Number of steps & 50 \\
Shift & True \\
Temporal reduction & 4 \\
Causal VAE & True \\
Guidance (T2V) & 7.5 \\
Guidance (I2V) & 3.0 \\
Text guidance oscillation & True \\
Image guidance oscillation & True \\
Temporal scale oscillation & True \\
Method & i2v\_head, t2v, t2i2v \\
Random seed & None \\
\hline
\end{tabular}

\label{tab:opensorav2}
\end{table}

We use the default Wan 2.2 GitHub implementation to generate ASMR videos with our ASMR prompt suite and ASMR image suite. The generation parameters are summarized in table~\ref{tab:wan}.
We evaluate three Wan models: T2V-A14B, I2V-A14B, and TI2V-5B.

\begin{table}[!t]
\centering
\small
\caption{Wan Sampling configuration used for video generation. The t2v-A14B, i2v-A14B, and ti2v-5B represent different video generation models of wan series under different settings.}
\begin{tabular}{l l}
\toprule
\textbf{Parameter} & \textbf{Value} \\
\midrule
task & t2v-A14B, i2v-A14B, ti2v-5B \\
size & 1280$\times$720 \\
frame\_num & None \\
ckpt\_dir & None \\
offload\_model & Ture \\
ulysses\_size & 1 \\
t5\_fsdp & True \\
t5\_cpu & True \\
dit\_fsdp & False \\
save\_file & None \\
prompt & None \\
use\_prompt\_extend & None \\
prompt\_extend\_method & local\_qwen \\
prompt\_extend\_model & None \\
prompt\_extend\_target\_lang & zh \\
base\_seed & -1 \\
image & None \\
sample\_solver & unipc \\
sample\_steps & None \\
sample\_shift & None \\
sample\_guide\_scale & None \\
convert\_model\_dtype & False \\
\midrule
src\_root\_path & None \\
refert\_num & 77 \\
replace\_flag & False \\
use\_relighting\_lora & False \\
\midrule
num\_clip & None \\
audio & None \\
enable\_tts & False \\
tts\_prompt\_audio & None \\
tts\_prompt\_text & None \\
tts\_text & None \\
pose\_video & None \\
start\_from\_ref & False \\
infer\_frames & 80 \\
\bottomrule
\end{tabular}

\label{tab:wan}
\end{table}

For other open-source video generation models, we adopt the DiffSynth-Studio GitHub implementation from the ModelScope community with acceleration enabled.
We evaluate the following video generation models: Hunyuan-I2V, Hunyuan-T2V, and StepFun-T2V.

For Veo3.1 and Sora2, we manually upload the images and prompts through their official web interfaces to obtain the generated videos. We adopt the default text–image–to–video generation settings, using text prompts from the ASMR prompt suite and starting frames from the ASMR image suite.
We also try the image-to-video settings for the Sora2.
For the watermark removal in the Sora2-generated videos, we also remove all watermarks for the ablation study.
We release all the Fake ASMR video datasets.

\section{Different prompt for \ours}
\label{app: different prompt}

We use the following prompt as the default prompt for the experiments in the main paper:

\subsection{Real--Fake Detection Prompt.}
We design a structured prompt as our default prompt for table~\ref{tab:v_gen} and table~\ref{tab: v under} that instructs the model to assess whether a given video is real or fake. 
The model is required to first generate an explicit reasoning process enclosed in \texttt{<think></think>} tags, followed by a binary decision enclosed in \texttt{<answer></answer>} tags, where ``1'' denotes a real video and ``0'' denotes a fake video.

\begin{AIbox}{Prompt for detecting real or fake.}
    Given the following video, please determine if the video is real or fake. First, think about the reasoning process, and then provide the answer. The reasoning process should be enclosed within \texttt{\textless think\textgreater \textless /think\textgreater} tags, and the answer should be enclosed within \textless answer\textgreater \textless /answer\textgreater tags. Respond with ``1'' for real and ``0'' for fake. For example: \textless think\textgreater reasoning process here \textless think\textgreater \textless answer\textgreater answer here \textless /answer\textgreater.   
\textless video \textgreater
\end{AIbox}

\subsection{Judgement Prompt without Reasoning.}
We design a concise judgment prompt that asks the model to directly classify a given video as real or fake without providing an explicit reasoning process. The model outputs a binary decision enclosed in \texttt{<answer></answer>} tags, where ``1'' denotes a real video and ``0'' denotes a fake one.
This is used in Table~\ref{tab:opensorav2}.

\begin{AIbox}{Prompt for \textit{Judgement} without reasoning}

Given the following video, please determine if the video is real or fake. 
Respond with ``1'' for real and ``0'' for fake. 
Provide the answer directly within \textless answer\textgreater \textless /answer\textgreater tags, like this: \textless answer\textgreater answer here \textless /answer\textgreater.

\textless video \textgreater
\end{AIbox}

\subsection{Preference Comparison Prompt with Reasoning.}
We design a preference comparison prompt in which the model is presented with two videos and asked to identify which one is real. The model first produces a brief reasoning process enclosed in \texttt{<think></think>} tags, followed by a decision enclosed in \texttt{<answer></answer>} tags, where ``1'' indicates that the first video is real and ``2'' indicates that the second video is real.
This is used in Table~\ref{tab: pair}.

\begin{AIbox}{Prompt for \textit{Preference Comparison} with reasoning}

\textless video\textgreater \textless video\textgreater
Based on the previous two videos, please identify which one is real. Respond with "1" if the first video is real and "2" if the second video is real. 
First, think about the reasoning process, and then provide the answer. The reasoning process should be enclosed within  \textless think\textgreater  \textless /think\textgreater tags, and the answer should be enclosed within \textless answer\textgreater \textless /answer\textgreater tags. Respond with "1" for real and "0" for fake. For example:  \textless think\textgreater reasoning process here  \textless / think\textgreater \textless answer\textgreater answer here \textless / answer\textgreater.

\end{AIbox}

\subsection{Preference Comparison Prompt without Reasoning.}
We further design a simplified preference comparison prompt that requires the model to directly select the real video from a pair without providing explicit reasoning. The model outputs its decision in \texttt{<answer></answer>} tags, where ``1'' and ``2'' denote the first and second videos, respectively.
This is used in Table~\ref{tab: pair}.

\begin{AIbox}{Prompt for \textit{Preference Comparison} without reasoning}

\textless video\textgreater \textless video\textgreater
Given the following video, please determine if the video is real or fake. Respond with "1" for real and "0" for fake. Provide the answer directly within \textless answer\textgreater \textless /answer\textgreater tags, like this: \textless answer\textgreater answer here \textless / answer\textgreater.

\end{AIbox}

\subsection{ASMR Prompt Suite Generation.}
We design a storyboard generation prompt that instructs the model to produce a single, coherent paragraph describing an ASMR video based on eight uniformly sampled frames. The prompt encourages integrated narration of the visual environment, object interactions, temporal dynamics, and implied auditory sensations, using concise and immersive language without explicit structural markers.
This is used in getting the ASMR Prompt Suite during the construction of \ours.

\begin{AIbox}{Prompt for getting storyboard of AMSR video}
Given 8 frames evenly sampled from an ASMR video, describe the overall scene in a single, continuous paragraph. 
Integrate information about the visual environment (background, lighting, textures, mood), the main subjects or objects, their actions and temporal dynamics, and the corresponding auditory sensations (\eg crisp cutting, gentle tapping, soft brushing, fluid pouring). 
Use concise yet immersive language that evokes sight, touch, and sound simultaneously. 
Avoid enumeration or structural markers—produce one coherent cinematic paragraph. \textless video\textgreater
\end{AIbox}

\section{Experiments}
\label{app: deiled exps}

\begin{table}[!t]
    \centering
    \caption{Impact of ASMR prompt suite on generated video realism. Scores are averaged across thinking/no-thinking conditions for each evaluation model. Our prompt suite universally lowers realism scores, assisting in ablation analysis.}
    \resizebox{0.6\linewidth}{!}{
    \begin{tabular}{l|cc|cc}
        \toprule
        \textbf{Evaluation} & \multicolumn{2}{c|}{\textbf{Wan2.2-A14B}} & \multicolumn{2}{c}{\textbf{Opensora-v2}} \\ 
        \textbf{Model} & w/o prompt & w/ prompt & w/o prompt & w/ prompt \\ 
        \midrule
        Qwen2.5-VL-7B & 5.1 & \textbf{0.0} & 8.1 & \textbf{1.0} \\
        GPT-4-turbo & 32.7 & \textbf{26.5} & 32.7 & \textbf{18.4} \\
        Gemini-2.5-pro & 52.1 & \textbf{33.7} & 72.5 & \textbf{52.1} \\
        \bottomrule
    \end{tabular}
    }
    \label{tab:prompt_ablation}
\end{table}

\begin{table}[!t]
    \centering
    \caption{Effect of reasoning strategy across video types. Scores for 'Generated Videos' are averaged across both subject models (Wan, Opensora) and prompt conditions (w/o, w/). Reasoning (\correctmark) consistently improves the perceived realism of actual videos but has mixed, model-dependent effects on generated content.}
    \resizebox{0.6\linewidth}{!}{
    \begin{tabular}{lc|c|c}
        \toprule
        \textbf{Evaluation Model} & \textbf{think?} & \textbf{Real Videos} & \textbf{Generated Videos (Avg)} \\ 
        \midrule
        \multirow{2}{*}{Qwen2.5-VL-7B} & \errormark & 98.0 & 4.6 \\
                                        & \correctmark & \textbf{100.0} & 2.6 \\ 
        \midrule
        \multirow{2}{*}{GPT-4-turbo}   & \errormark & 73.5 & 40.3 \\
                                        & \correctmark & \textbf{89.8} & \textbf{14.8} \\ 
        \midrule
        \multirow{2}{*}{Gemini-2.5-pro} & \errormark & 71.4 & 51.6 \\
                                        & \correctmark & \textbf{75.5} & 53.6 \\ 
        \bottomrule
    \end{tabular}
    }
    \label{tab:reasoning_ablation}
\end{table}

\begin{table}[!t]
    \centering
    \caption{Comparison of evaluation model strictness. Realism scores are averaged across all reasoning conditions, subject models, and prompt suites. Qwen2.5-VL-7B emerges as the most critical judge of generated videos.}
    \resizebox{0.5\linewidth}{!}{
    \begin{tabular}{l|c}
        \toprule
        \textbf{Evaluation Model} & \textbf{Avg. Judgement Score} \\
        & \textbf{(Across Generated Videos)} \\
        \midrule
        \textbf{Qwen2.5-VL-7B}  & \textbf{3.6} \\
        GPT-4-turbo              & 27.5 \\
        Gemini-2.5-pro           & 52.6 \\
        \bottomrule
    \end{tabular}
    }
    \label{tab:evaluator_comparison}
\end{table}

\begin{table}[!t]
    \centering
    \caption{Ablation on reasoning strategies and ASMR prompt suite. Answer with thinking \correctmark and without thinking \errormark; generate videos with our ASMR prompt suite and without our prompt.}
    \resizebox{0.9\linewidth}{!}{
    \begin{tabular}{ccccccc}
        \toprule
        \multirow{2}{*}{\textbf{Eval}}& \multirow{2}*{think?} & \multirow{2}*{real} & \multicolumn{2}{c}{\textbf{Wan2.2-A14B}} & \multicolumn{2}{c}{\textbf{Opensora-v2}} \\ 
        \cmidrule(lr){4-5} \cmidrule(lr){6-7} 
          & & & w/o prompt & w/ prompt & w/o prompt & w/ prompt \\ 
        \hline
        \rowcolor{mygray}\multicolumn{7}{c}{\emph{Qwen2.5-VL-7B}} \\
        \hline
        \multirow{2}*{Judgement} & \errormark & 98.0 & 6.1 & 0.0 & 10.2 & 2.0  \\
         & \correctmark & 100.0 & 4.1 & 0.0 & 6.1 & 0  \\ 
         \hline
        \rowcolor{mygray}\multicolumn{7}{c}{\emph{GPT-4-turbo}} \\
        \hline
        \multirow{2}*{Judgement}  & \errormark & 73.5 & 49.0 & 40.8 & 42.9 &  28.6  \\
         & \correctmark & 89.8 & 16.3 & 12.2 & 22.4 & 8.2 \\ 
         \hline
        \rowcolor{mygray}\multicolumn{7}{c}{\emph{Gemini-2.5-pro}} \\
        \hline
        \multirow{2}*{Judgement}  & \errormark & 71.4 & 51.0 & 32.7 & 69.4 & 53.1  \\ 
         & \correctmark & 75.5 & 53.1 & 34.7 & 75.5 & 51.0  \\ 
        \bottomrule
    \end{tabular}
    }
    \label{tab:accuracy_comparison}
\end{table}

\begin{table}[!t]
    \centering
    \caption{The Impact of Reasoning (\correctmark) on Evaluation Stability. scores are averaged across prompt conditions. Reasoning consistently improves accuracy for Opensora-v2 and remarkably stabilizes scores to an identical level for Wan2.2-A14B across both 'Pair' and 'Pair shuffle' modes.}
    \resizebox{0.7\linewidth}{!}{
    \begin{tabular}{l c cc cc}
        \toprule
        \multirow{2}{*}{\textbf{Evaluation Mode}} & \multirow{2}*{think?} & \multicolumn{2}{c}{\textbf{Wan2.2-A14B (Avg)}} & \multicolumn{2}{c}{\textbf{Opensora-v2 (Avg)}} \\ 
        \cmidrule(lr){3-4} \cmidrule(lr){5-6}
        & & w/o think & w/ think & w/o think & w/ think \\
        \midrule
        Pair          & \errormark vs \correctmark & 45.85 & \textbf{56.2} & 55.2 & \textbf{72.95} \\
        Pair shuffle & \errormark vs \correctmark & 55.2 & \textbf{56.2} & 52.1 & \textbf{64.6} \\ 
        \bottomrule
    \end{tabular}
    }
    \label{tab:reasoning_stability}
\end{table}

\begin{table}[!t]
    \centering
    \caption{Sensitivity of Evaluation to Pairing Type (No Thinking). Without reasoning (\errormark), 'Pair shuffle' sometimes yields significantly higher or lower scores than 'Pair' mode, contradicting the assumption that aligned pairing is always easier/higher scoring. Model-dependent volatility is prominent in simple preference tasks.}
    \resizebox{0.9\linewidth}{!}{
    \begin{tabular}{l c cc cc}
        \toprule
        \multirow{2}{*}{\textbf{Evaluation Mode}} & \multirow{2}*{think?} & \multicolumn{2}{c}{\textbf{Wan2.2-A14B}} & \multicolumn{2}{c}{\textbf{Opensora-v2}} \\ 
        \cmidrule(lr){3-4} \cmidrule(lr){5-6}
        & & w/o prompt & w/ prompt & w/o prompt & w/ prompt \\
        \midrule
        \textbf{Pair}          & \errormark & 43.8 & 47.9 & 58.3 & 52.1 \\
        \textbf{Pair shuffle} & \errormark & 58.3 & 52.1 & 50.0 & 54.2 \\ 
        \bottomrule
    \end{tabular}
    }
    \label{tab:pairing_sensitivity}
\end{table}

\begin{table}[!t]
    \centering
    \caption{Ablation on Preference versus judgement. Answer with thinking \correctmark and without thinking \errormark; generate videos with our ASMR prompt suite and without our prompt. Eval on Qwen2.5-VL-7B. Preference evaluation is easier than judgment.
    }
    \resizebox{0.9\linewidth}{!}{
    \begin{tabular}{cccccc}
        \toprule
        \multirow{2}{*}{\textbf{Eval}}& \multirow{2}*{think?} & \multicolumn{2}{c}{\textbf{Wan2.2-A14B}} & \multicolumn{2}{c}{\textbf{Opensora-v2}} \\ 
        \cmidrule(lr){3-4} \cmidrule(lr){5-6} 
           & & w/o prompt & w/ prompt & w/o prompt & w/ prompt  \\ 
        \hline

         Pair & \errormark &43.8 & 47.9 & 58.3 & 52.1  \\
         Pair shuffle & \errormark & 58.3 & 52.1 & 50.0 & 54.2 \\

        \hline
          Pair & \correctmark &  56.2 &	56.2&	77.1	& 68.8 \\
         Pair shuffle  & \correctmark &  56.2 & 56.2 & 62.5 & 66.7 \\
        
        \bottomrule
    \end{tabular}
    }
    \label{tab: pair}
\end{table}

\subsection{Detection Results}

Tab.~\ref{tab: v under} reports a comprehensive evaluation of 11 VLMs on \ours, along with human and random baselines as the upper and lower bounds. 
We use 7 state-of-the-art VGMs to generate fake ASMR videos.
Overall, proprietary VLMs show stronger performance than most open-source models, with Gemini-3-pro achieving the best results and Gemini-2.5-pro following closely.
However, their ability to reliably distinguish real from fake videos remains limited: the best average accuracy across all VGM-generated fake videos is only 76.27, about 13 points below the human score of 89.11. 
Most open-source models perform poorly, indicating substantial room for improvement.

Moreover, we observe a potential “intra-family bias” phenomenon, where evaluation models appear particularly aligned with video generators from the same ecosystem (\eg Gemini with Veo, GPT with Sora). 
Specifically, video understanding models tend to assign anomalously low detection accuracy to videos generated by their corresponding proprietary video generation models compared to other open-source generators. 
For example, the Gemini series exhibits unusually low scores on Veo-generated videos relative to its evaluations of other video generation models. 
We hypothesize that this behavior may stem from the training pipeline of these video generation systems, where the corresponding video understanding models are potentially used as reward models or evaluators during optimization.

\input{table/v_u}

\begin{figure*}[!h]
    \centering
    \includegraphics[width=\linewidth]{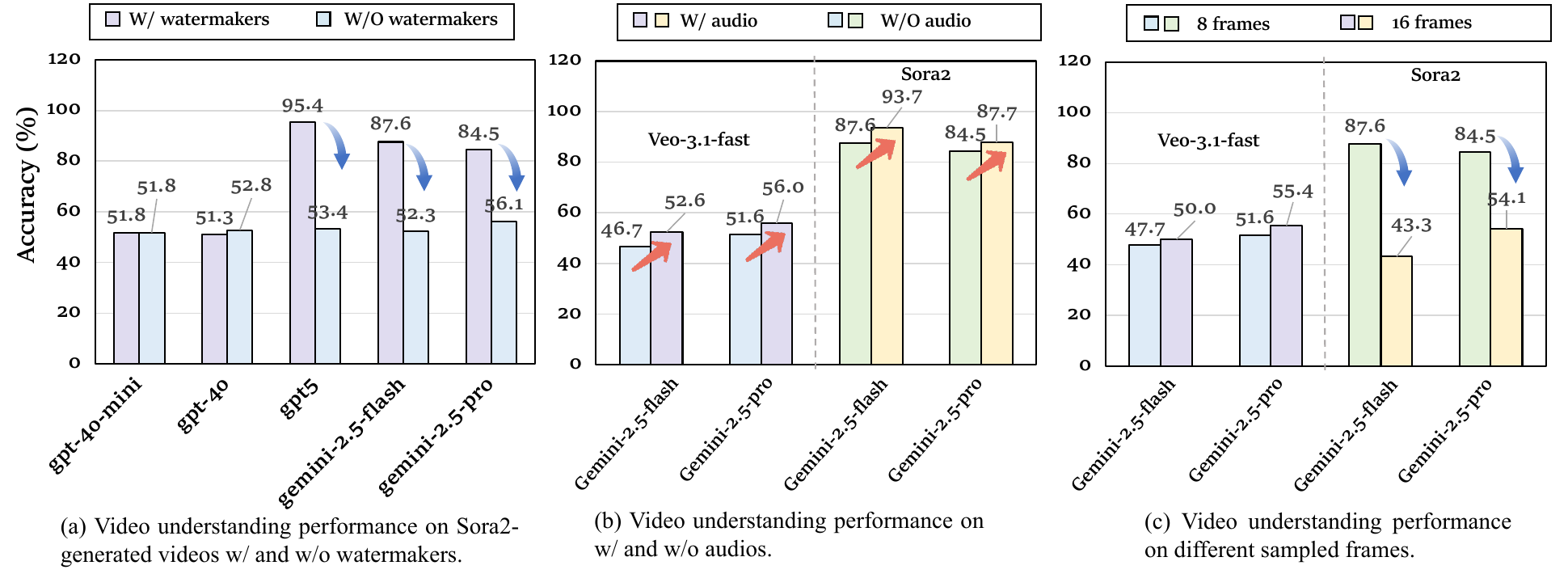}
    \caption{Ablation study. (a) VLMs’ performance drops after the sora watermark removal. (b) Incorporating audio along with
visual inputs generally improves reality detection accuracy. (c) More frames as the input degrade the performance of the detection by the VLMs.}
    \label{fig:app abla}
\end{figure*}



\begin{table*}[h!]
\centering
\caption{Prediction statistics of Qwen3-VL-8B-Instruct on real and generated videos.}
\begin{tabular}{lccc}
\hline
\rowcolor{gray!15}
\textbf{Source (real/Video Gen Model)} & \textbf{Valid Total Answer Number} & \textbf{Predict = Fake} & \textbf{Predict = Real} \\
\hline
Real & 100 & 23 & 77 \\
\hline
Veo3.1-fast & 99 &	37	& 62 \\
Sora2 & 95 & 94 & 1 \\
Hunyuan & 100 & 32 & 68 \\
Opensora (i2v) & 100 & 26 & 74 \\
Wan (it2v) & 99 & 35 & 64 \\
Wan (i2v) & 89 & 28 & 61 \\
StepVideo (t2v) & 98 & 89 & 9 \\
\hline
\end{tabular}
\label{tab:qwen3_3b_pred}
\end{table*}

\begin{table*}[h!]
\centering
\caption{Prediction statistics of Qwen3-vl-235b-Instruct on real and generated videos.}
\begin{tabular}{lccc}
\hline
\rowcolor{gray!15}
\textbf{Source (real/Video Gen Model)} & \textbf{Valid Total Answer Number} & \textbf{Predict = Fake} & \textbf{Predict = Real} \\
\hline
Real & 95 & 7 & 88 \\
\hline
Veo3.1-fast & 96 &	19 &	77 \\
Sora2 & 92 & 63 & 29 \\
Hunyuan & 98 & 5 & 93 \\
Opensora (i2v) & 94 & 8 & 86 \\
Wan (it2v) & 93 & 11 & 82 \\
Wan (i2v) & 85 & 9 & 76 \\
StepVideo (t2v) & 95 & 84 & 11 \\
\hline
\end{tabular}
\label{tab:qwen3_235b_pred}
\end{table*}

\begin{table*}[h!]
\centering
\caption{Prediction statistics of Qwen2.5-vl-72B-Instruct on real and generated videos.}
\begin{tabular}{lccc}
\hline
\rowcolor{gray!15}
\textbf{Source (real/Video Gen Model)} & \textbf{Valid Total Answer Number} & \textbf{Predict = Fake} & \textbf{Predict = Real} \\
\hline
Real Videos & 95 & 7 & 88 \\
\hline
Veo3.1-fast & 100 &	17	& 83 \\
Sora2 (IT2V) & 92 & 63 & 29 \\
Hunyuan (IT2V) & 98 & 5 & 93 \\
OpenSora (I2V) & 94 & 8 & 86 \\
Wan (IT2V) & 93 & 11 & 82 \\
Wan (I2V) & 85 & 9 & 76 \\
StepVideo (T2V) & 95 & 84 & 11 \\
\hline
\end{tabular}
\label{tab:qwen2.5_72b_pred}
\end{table*}

\begin{table*}[h!]
\centering
\caption{Prediction statistics of GPT5 on real and generated videos.}
\begin{tabular}{lccc}
\hline
\rowcolor{gray!15}
\textbf{Source (real/Video Gen Model)} & \textbf{Valid Total Answer Number} & \textbf{Predict = Fake} & \textbf{Predict = Real} \\
\hline
Real Videos & 100 & 2 & 98 \\
\hline
Veo3.1-fast & 99	& 11 &	88 \\
Sora2 (IT2V) & 97 & 90 & 7 \\
Hunyuan (IT2V) & 100 & 15 & 85 \\
OpenSora (T2V) & 99 & 16 & 83 \\
OpenSora (I2V) & 99 & 15 & 84 \\
Wan (IT2V) & 100 & 17 & 83 \\
Wan (I2V) & 90 & 7 & 83 \\
StepVideo (T2V) & 99 & 89 & 10 \\
\hline
\end{tabular}
\label{tab:GPT5_pred}
\end{table*}

\begin{table*}[h!]
\centering
\caption{Prediction statistics of Qwen3-VL-30B on real and generated videos.}
\begin{tabular}{lccc}
\hline
\rowcolor{gray!15}
\textbf{Source (real/Video Gen Model)} & \textbf{Valid Total Answer Number} & \textbf{Predict = Fake} & \textbf{Predict = Real} \\
\hline
Real Videos & 94 & 19 & 75 \\
\hline
Veo3.1-fast & 90&	20	& 70 \\
Sora2 (IT2V) & 91 & 20 & 71 \\
Hunyuan (IT2V) & 95 & 19 & 76 \\
OpenSora (I2V) & 95 & 14 & 81 \\
Wan (IT2V) & 96 & 29 & 67 \\
Wan (I2V) & 86 & 14 & 72 \\
StepVideo (T2V) & 97 & 81 & 16 \\
\hline
\end{tabular}
\label{tab:qwen3-vl-30b_pred}
\end{table*}

\begin{table*}[h!]
\centering
\caption{Prediction statistics of GLM-4V on real and generated videos.}
\begin{tabular}{lccc}
\hline
\rowcolor{gray!15}
\textbf{Source (real/Video Gen Model)} & \textbf{Valid Total Answer Number} & \textbf{Predict = Fake} & \textbf{Predict = Real} \\
\hline
Real Videos & 80 & 15 & 65 \\
\hline
Veo3.1-fast & 77&	20	&57 \\
Sora2 (IT2V) & 80 & 37 & 43 \\
Hunyuan (IT2V) & 79 & 32 & 47 \\
OpenSora (I2V) & 77 & 39 & 38 \\
Wan (IT2V) & 78 & 26 & 52 \\
Wan (I2V) & 73 & 19 & 54 \\
StepVideo (T2V) & 91 & 84 & 7 \\
\hline
\end{tabular}
\label{tab:glm4v_pred}
\end{table*}

\begin{table*}[h!]
\centering
\caption{Prediction statistics of Gemini-2.5-pro on real and generated videos.}
\begin{tabular}{lccc}
\hline
\rowcolor{gray!15}
\textbf{Source (real/Video Gen Model)} & \textbf{Valid Total Answer Number} & \textbf{Predict = Fake} & \textbf{Predict = Real} \\
\hline
Real Videos & 92 & 10 & 82 \\
\hline
Veo3.1-fast & 99 &	16 &	83 \\
Sora2 IT2V & 95 & 76 & 19 \\
Sora2 T2V & 96 & 93 & 3 \\
Hunyuan T2V & 97 & 18 & 79 \\
Hunyuan IT2V & 92 & 39 & 53 \\
OpenSora T2I2V & 96 & 42 & 54 \\
OpenSora T2V & 97 & 38 & 59 \\
OpenSora I2V & 91 & 32 & 59 \\
Wan I2V & 84 & 22 & 62 \\
Wan IT2V & 99 & 33 & 66 \\
Wan T2V & 92 & 20 & 72 \\
StepVideo T2V & 91 & 79 & 12 \\
\hline
\end{tabular}
\label{tab:gemini2.5-pro_pred}
\end{table*}

\begin{table*}[h!]
\centering
\caption{Prediction statistics of Gemini-2.5-flash on real and generated videos.}
\begin{tabular}{lccc}
\hline
\rowcolor{gray!15}
\textbf{Source (real/Video Gen Model)} & \textbf{Valid Total Answer Number} & \textbf{Predict = Fake} & \textbf{Predict = Real} \\
\hline
Real Videos & 98 & 20 & 78 \\
\hline
veo3.1-fast & 100 &	5	& 95\\
Sora2 IT2V & 95 & 91 & 4 \\
Sora2 T2V & 96 & 96 & 0 \\
Hunyuan IT2V & 98 & 26 & 72 \\
Hunyuan T2V & 98 & 25 & 73 \\
OpenSora T2I2V & 97 & 44 & 53 \\
OpenSora T2V & 94 & 27 & 67 \\
OpenSora I2V & 96 & 29 & 67 \\
Wan IT2V & 95 & 29 & 66 \\
Wan T2V & 94 & 23 & 71 \\
Wan I2V & 85 & 20 & 65 \\
\hline
\end{tabular}
\label{tab:gemini2.5-flash_pred}
\end{table*}

\begin{table*}[h!]
\centering
\caption{Prediction statistics of GPT-4o on real and generated videos.}
\begin{tabular}{lccc}
\hline
\rowcolor{gray!15}
\textbf{Source (real/Video Gen Model)} & \textbf{Valid Total Answer Number} & \textbf{Predict = Fake} & \textbf{Predict = Real} \\
\hline
Real Videos & 100 & 2 & 98 \\
\hline
veo3.1-fast & 100	& 11	& 89\\
Sora2 IT2V & 97 & 3 & 94 \\
Sora2 T2V & 100 & 9 & 91 \\
Hunyuan IT2V & 100 & 15 & 85 \\
Hunyuan T2V & 100 & 7 & 93 \\
OpenSora T2I2V & 100 & 18 & 82 \\
OpenSora T2V & 100 & 10 & 90 \\
OpenSora I2V & 100 & 15 & 85 \\
Wan IT2V & 100 & 13 & 87 \\
Wan T2V & 100 & 13 & 87 \\
Wan I2V & 90 & 7 & 83 \\
StepVideo T2V & 100 & 92 & 8 \\
\hline
\end{tabular}
\label{tab:gpt-4o_pred}
\end{table*}

\begin{table*}[h!]
\centering
\caption{Prediction statistics of GPT-4o-mini on real and generated videos.}
\begin{tabular}{lccc}
\hline
\rowcolor{gray!15}
\textbf{Source (real/Video Gen Model)} & \textbf{Valid Total Answer Number} & \textbf{Predict = Fake} & \textbf{Predict = Real} \\
\hline
Real Videos & 100 & 6 & 94 \\
\hline
veo3.1-fast & 100& 17 & 83 \\
Sora2 IT2V & 97 & 8 & 89 \\
Sora2 T2V & 100 & 16 & 84 \\
Hunyuan IT2V & 100 & 7 & 93 \\
Hunyuan T2V & 100 & 8 & 92 \\
OpenSora T2I2V & 100 & 14 & 86 \\
OpenSora T2V & 100 & 14 & 86 \\
OpenSora I2V & 100 & 12 & 88 \\
Wan IT2V & 100 & 7 & 93 \\
Wan T2V & 100 & 12 & 88 \\
Wan I2V & 90 & 8 & 82 \\
StepVideo T2V & 100 & 84 & 16 \\
\hline
\end{tabular}
\label{tab:gpt-4o-mini_pred}
\end{table*}


\subsection{Real-Fake Evaluation Results}

Tables~\ref{tab:qwen3_3b_pred}-\ref{tab:gpt-4o-mini_pred} are the hard-level reality evaluation results across mainstream video understanding models and generation models.

%% file: table/v_u.tex
\begin{table*}[t]
    \centering
    \caption{
    \textbf{Detection Accuracy for Video Understanding.}
    This table compares the performance of SOTA VLMs using \ours. 
    A higher score indicates better performance. 
\colorbox{gold}{\textcolor{black}{Gold}}, \colorbox{silver}{\textcolor{black}{silver}}, and 
\colorbox{bronze}{\textcolor{black}{bronze}} denote the best, second, and third performers.}   
\resizebox{\linewidth}{!}{
    \begin{tabular}{lcc|ccccc|cc}
 
\toprule
      Model & Veo3.1-Fast  & Sora2  & Wan2.2-A14B & Wan2.2-5B & Opensora-V2 & HunyuanVideo & StepVideo & \textbf{Avg. ($\uparrow$)} & Rank\\

      \hline
      Random & 50 &50 & 50  & 50 & 50 & 50 & 50 & 50 & 13\\
      Human  & 81.25 & 91.25 & 86.25 & 91.25 & 91.25 & 91.25 & 91.25 & 89.11 & \cellcolor{gold}{1}  \\
      \hline
        \rowcolor{mygray}\multicolumn{10}{c}{\emph{Open-source Models}  } \\
        \hline
       
      Qwen3-VL-8B & 57.79 & 87.69 & 55.56 & 56.28 & 51.50 & 54.50 & 83.84  & 63.88 & 5\\
      Qwen3-VL-30B-A3B & 51.08 & 51.35 & 49.44 & 54.74 & 47.09 & 49.74 & 81.68 & 54.87 & 12\\
        Qwen2.5-VL-72B & 49.50 & 71.07	&51.05	&51.00&	54.50	&53.50	&81.50 & 58.87 & 10 \\
        Qwen3-VL-235B-A22B & 56.53 & 80.75	&53.89	&52.66	&50.79	&48.19	&90.53 & 61.91 & 8\\
            GLM-4.5V & 54.64 &63.75 & 54.90 & 57.59 & 66.24 & 61.01 & 87.13 & 63.61 & 6\\
        \hline
        \rowcolor{mygray}\multicolumn{10}{c}{\emph{Proprietary Models}} \\
        \hline
        GPT-4o-mini  & 52.50     & 51.78 & 53.68 & 50.50 & 53.00 & 50.50 & 89.00  & 57.28 & 11\\ 
        GPT-4o     & 51.50        & 51.27 & 55.26 & 55.50 & 56.50 & 56.50 & 95.00 & 60.22 & 9\\ 
        GPT-5 & 54.55 & 95.43 &	55.26 &	57.50 &	56.78 &	56.50 &	93.97 & 67.14 & 4\\
        Gemini-2.5-Flash & 47.72  & 87.56 & 53.55 & 55.44 & 55.15 & 53.06 & 78.63 & 61.59 & 7\\ 
        \rowcolor{lightpurple}
        + Audio & 52.55 & 93.65 & 53.55 & 55.44 & 55.15 & 53.06 & 78.63   &  63.15 & 8\\ 
        Gemini-2.5-Pro   & 51.56  & 84.49 & 59.09 & 60.21 & 62.30 & 65.76 & 87.98 & 67.34 & \cellcolor{bronze}{3}\\ 
        \rowcolor{lightpurple}
        + Audio & 56.00 &	87.72 & 59.09 & 60.21 & 62.30 & 65.76 & 87.98  & 68.44 & \cellcolor{bronze}{3}\\ 
         Gemini-3-Pro-Preview   & 77.89 & 89.90 & 57.67 & 73.87 & 65.83 & 80.90 & 87.94 & 76.27 & \cellcolor{silver}{2} \\
    \bottomrule
    \end{tabular}
    }
    \label{tab: v under}
\end{table*}
  

      %
      %

      